\documentclass[10pt,twocolumn,letterpaper]{article}
\usepackage[pagenumbers]{cvpr}

\usepackage{graphicx}
\usepackage{amsmath}
\usepackage{amssymb}
\usepackage{booktabs}
\usepackage{xcolor}
\usepackage{colortbl}
\usepackage{times}
\usepackage{epsfig}
\usepackage{soul}
\usepackage{caption}
\usepackage{bbold}
\usepackage[mathletters]{ucs}
\usepackage[utf8x]{inputenc}

\renewcommand{\paragraph}[1]{\vspace{1.25mm}\noindent\textbf{#1}}
\makeatletter\renewcommand\paragraph{\@startsection{paragraph}{4}{\z@} {.5em \@plus1ex \@minus.2ex}{-.5em}{\normalfont\normalsize\bfseries}}\makeatother

\definecolor{convcolor}{HTML}{412F8A}
\definecolor{vitcolor}{HTML}{fc8e62}
\definecolor{swincolor}{HTML}{fc6562}
\definecolor{pa}{HTML}{C209C1}
\colorlet{dilatedcolor}{purple}
\usepackage[pagebackref,breaklinks,colorlinks,citecolor=pa,urlcolor=blue]{hyperref}

\usepackage[capitalize]{cleveref}
\crefname{section}{Sec.}{Secs.}
\Crefname{section}{Section}{Sections}
\Crefname{table}{Table}{Tables}
\crefname{table}{Tab.}{Tabs.}
\Crefname{appendix}{Appendix}{Appendices}

\def\cvprPaperID{xxxxx}
\def\confName{CVPR}
\def\confYear{2022}

\DeclareMathAlphabet\mathbfcal{OMS}{cmsy}{b}{n}
\newcommand{\natten}{$\mathcal{N}ATTEN$}

\newcommand{\sq}{\textsuperscript{2}}
\newcommand{\ips}{\tiny{imgs/sec}}
\newcommand{\fps}{\tiny{fps}}
\newcommand{\strr}{$^\star$}
\newcommand{\dgr}{$^\dag$}
\newcommand{\dgg}{$^\ddagger$}
\newcommand{\iso}{~\textit{(iso.)}}
\newcommand{\pls}{\textsuperscript{+}}
\newcommand{\bigO}{\mathcal{O}}

\newcommand{\convcolor}[1]{\textcolor{convcolor}{#1}}
\newcommand{\vitcolor}[1]{\textcolor{vitcolor}{#1}}
\newcommand{\swincolor}[1]{\textcolor{swincolor}{#1}}
\newcommand{\natcolor}[1]{\textcolor{pa}{#1}}

\newcommand{\wb}{\swincolor{$\mathbf{\circ}$\,}} 
\newcommand{\vb}{\vitcolor{$\bullet$\,}} 
\newcommand{\cb}{\convcolor{$\bullet$\,}} 
\newcommand{\nb}{\natcolor{$\mathbf{\circ}$\,}}
\newcommand{\db}{\natcolor{$\bullet$\,}}

\newcommand{\grayrow}{\rowcolor[gray]{.95}}
\newcommand{\ours}{\grayrow\db}
\newcommand{\oursn}{\grayrow\nb}

\newcommand{\imagenetnotes}{Throughput and peak memory usage are measured from forward passes with a batch size of 256 on a single NVIDIA A100 GPU.}
\newcommand{\downstreamnotes}{Throughput is measured on a single NVIDIA A100 GPU.}
\newcommand{\isonotes}{To compare self attention and NA/DiNA fairly, we ran ViT\pls, which uses relative positional biases in attention layers, instead of the one-time absolute positional encoding in the original ViT. }

\begin{document}
\title{
    \textcolor{dilatedcolor}{$\mathbb{Dilated}$} Neighborhood Attention Transformer
}
\author{
Ali Hassani\textsuperscript{1}, Humphrey Shi\textsuperscript{1,2} \\
{\small  \textsuperscript{1}SHI Lab @ U of Oregon \& UIUC, \textsuperscript{2}Picsart AI Research (PAIR)}\\
}

\twocolumn[{
\maketitle
\begin{center}
    \includegraphics[width=0.99\textwidth]{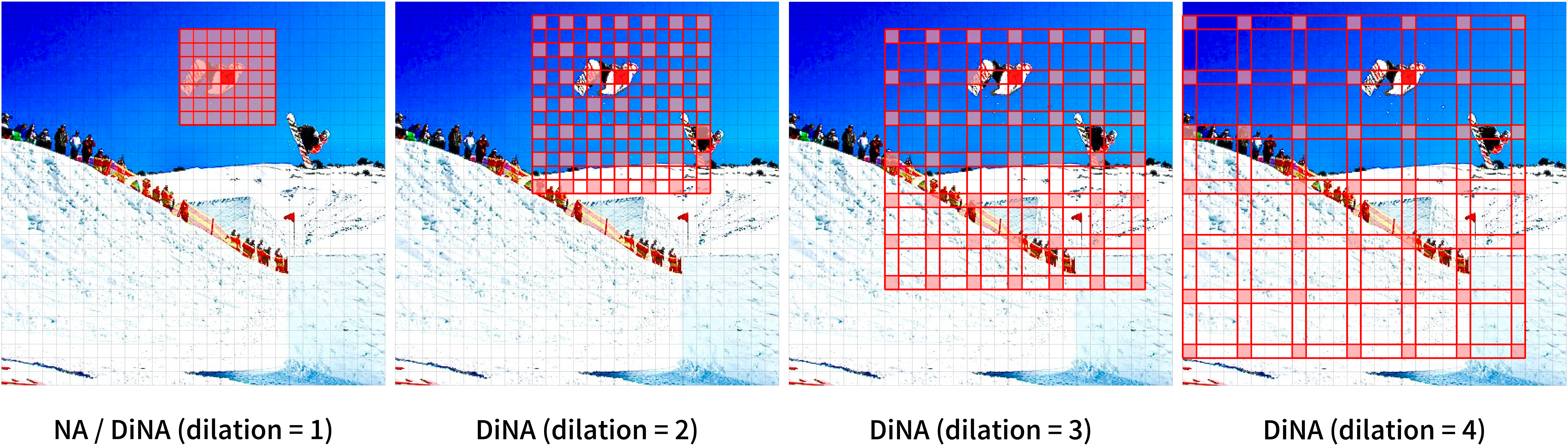}
    \captionsetup{type=figure}
    \captionof{figure}{
    An illustration of a single pixel's attention span in Neighborhood Attention (NA) and Dilated Neighborhood Attention (DiNA).
    NA localizes attention to the pixel's nearest neighbors.
    DiNA extends NA's {\textcolor{blue}{\textit{\textbf{local attention}}}} to a less constrained {\textcolor{dilatedcolor}{\textit{\textbf{sparse global attention}}}} without additional computational burden.
    Transformers comprised of both NA and DiNA are capable of preserving locality, maintaining translational equivariance, expanding the receptive field exponentially, and capturing longer-range inter-dependencies, leading to significant performance boosts in downstream vision tasks.
    }
    \label{fig:introfig}
\end{center}
}]

\thispagestyle{empty}

\begin{abstract}
Transformers are quickly becoming one of the most heavily applied deep learning architectures across modalities, domains, and tasks.
In vision, on top of ongoing efforts into plain transformers, hierarchical transformers have also gained significant attention, thanks to their performance and easy integration into existing frameworks.
These models typically employ localized attention mechanisms, such as the sliding-window Neighborhood Attention (NA) or Swin Transformer's Shifted Window Self Attention.
While effective at reducing self attention's quadratic complexity, local attention weakens two of the most desirable properties of self attention: long range inter-dependency modeling, and global receptive field.
In this paper, we introduce \textbf{Dilated Neighborhood Attention (DiNA)}, a natural, flexible and efficient extension to NA that can capture more global context and expand receptive fields exponentially \textbf{at no additional cost}. 
\textcolor{blue}{\textbf{NA's local attention}} and \textcolor{dilatedcolor}{\textbf{DiNA's sparse global attention}} complement each other, and therefore we introduce \textbf{Dilated Neighborhood Attention Transformer (DiNAT)}, a new hierarchical vision transformer built upon both.
DiNAT variants enjoy significant improvements over strong baselines such as NAT, Swin, and ConvNeXt.
Our large model is faster and ahead of its Swin counterpart by 1.6\% box AP in COCO object detection, 1.4\% mask AP in COCO instance segmentation, and 1.4\% mIoU in ADE20K semantic segmentation. 
Paired with new frameworks, our large variant is the new state of the art panoptic segmentation model on COCO (58.5 PQ) and ADE20K (49.4 PQ), and instance segmentation model on Cityscapes (45.1 AP) and ADE20K (35.4 AP) (no extra data).
It also matches the state of the art specialized semantic segmentation models on ADE20K (58.1 mIoU), and ranks second on Cityscapes (84.5 mIoU) (no extra data).
To support and encourage research in this direction, in vision and beyond, we open-source our project at: \url{https://github.com/SHI-Labs/Neighborhood-Attention-Transformer}.
\end{abstract}

\section{Introduction}
Transformers~\cite{vaswani2017attention} have made a significant contribution to AI research, starting with natural language understanding~\cite{radford2018improving,devlin2019bert} before being applied to other modalities such as speech~\cite{gulati2020conformer} and vision~\cite{parmar2018image,dosovitskiy2020image}, thanks to their universal architecture built upon self attention. This success inspired efforts into attention-based models in vision, from backbone networks~\cite{ramachandran2019stand,vaswani2021scaling}, to more specific applications including image generation and density modeling~\cite{parmar2018image,child2019generating}, object detection~\cite{carion2020end}, image segmentation~\cite{huang2019ccnet,wang2020axial}, and more.

Vision Transformer (ViT)~\cite{dosovitskiy2020image} was one of the first major demonstrations of transformers as direct alternatives to Convolutional Neural Networks (CNNs)~\cite{lecun1989backpropagation,krizhevsky2012imagenet,he2016deep}, the de facto standard in vision.
ViT treats an image as a sequence of patches and uses a plain transformer encoder to encode and classify images. It demonstrated competitive performance to CNNs on large scale image classification, and resulted in a surge in vision research focused on transformer-based architectures as competitors to CNNs~\cite{touvron2020training,touvron2021going}.

\begin{figure}[t]
    \centering
    \includegraphics[width=0.47\textwidth]{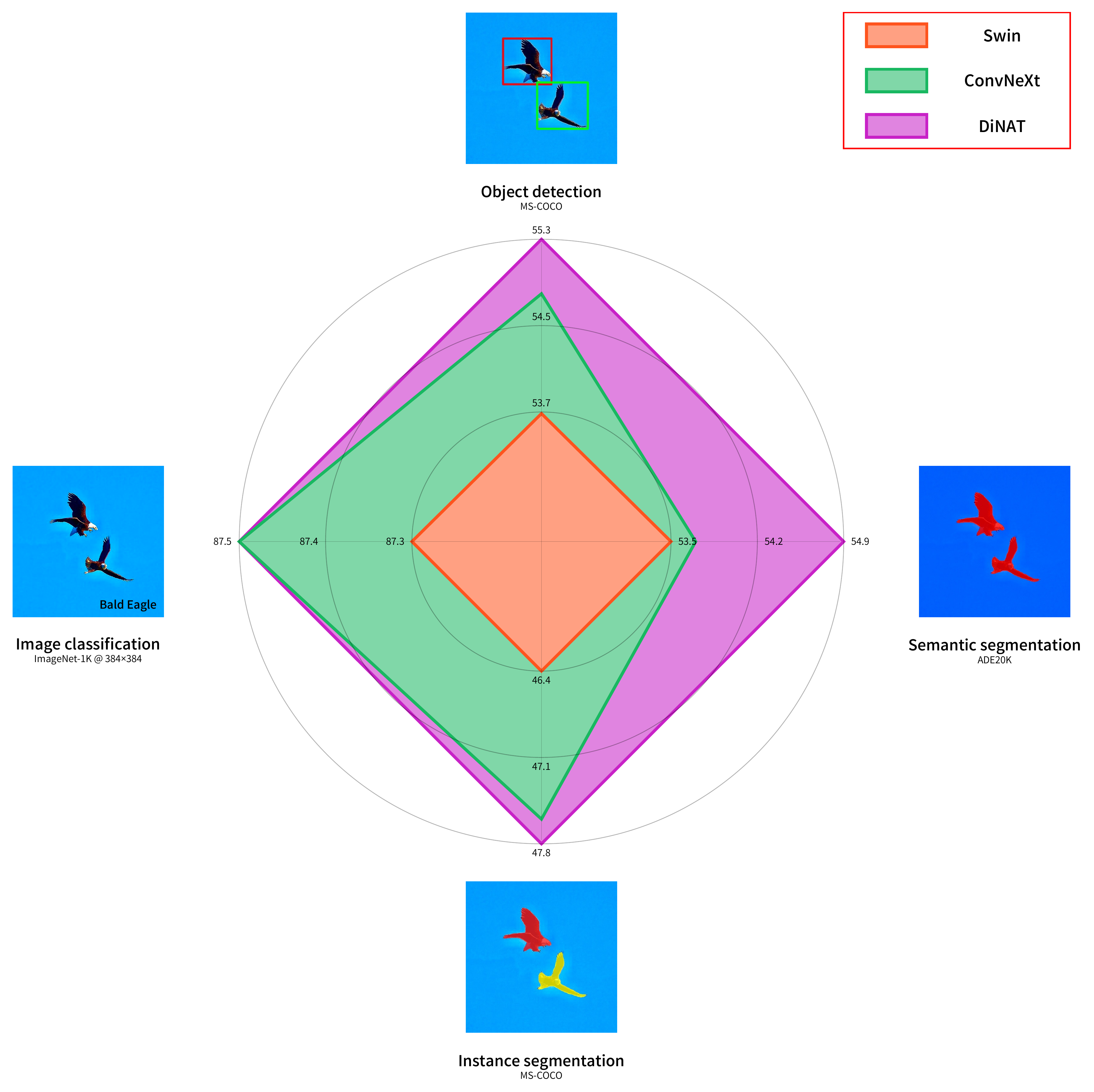}
    \caption{
    \textbf{Radar chart comparing Swin-L, ConvNeXt-L, and our DiNAT-L across various visual recognition tasks.}
    Both ConvNeXt and our DiNAT surpass Swin on all tasks. 
    DiNAT further exhibits noticeable improvements over downstream dense recognition tasks compared to ConvNeXt.
    }
    \label{fig:radarchart}
\end{figure}

Vision transformers and CNNs are different not only in terms of architecture and building blocks, but also in how they treat data.
CNNs typically downsample inputs gradually as they pass through the model and construct hierarchical feature maps. This hierarchical design is crucial for vision, as objects vary in scale, and high-resolution feature maps are important to dense tasks, such as segmentation. On the other hand, transformers are known for their fixed dimensionality throughout the model, and as a result, plain ViTs downsample inputs aggressively from the very beginning to alleviate the quadratic cost of self attention, which in turn hinders the application of plain ViTs as backbones to dense vision tasks. 

While research in applying plain ViTs to dense vision tasks continues~\cite{li2022exploring,he2022masked}, research into \textit{\textbf{hierarchical vision transformers}} quickly became dominant~\cite{wang2021pyramid,liu2021swin} and continues to grow~\cite{liu2022swin,hassani2022neighborhood}.
A key advantage of these hierarchical transformer models is their ease of integration with existing hierarchical vision frameworks.
Inspired by existing CNNs, hierarchical vision transformers are comprised of multiple (typically 4) levels of transformer encoders, with downsampling modules in between, and a less aggressive initial downsampling (i.e. $1/4$ instead of $1/16$). Earlier layers in hierarchical transformers, if using unrestricted self attention, would bear the same quadratically growing complexity and memory usage with respect to input resolution, making them intractable for higher resolution images. Therefore, hierarchical transformers typically employ certain \textit{\textbf{local attention}} mechanisms.

Swin Transformer~\cite{liu2021swin}, one of the earliest hierarchical vision transformers, utilizes a Window Self Attention (WSA) module, followed by a pixel-shifted Window Self Attention (SWSA), both of which localize self attention to non-overlapping sub-windows. 
This reduces the cost of self attention, making its time and space complexity linear with respect to resolution.
SWSA is identical to WSA, but with a shift in feature map pixels preceding it, and followed by a reverse shift. This is essential to its performance, as it allows out-of-window interactions, and therefore the expansion of its receptive field.
One of the major advantages of Swin is efficiency, as pixel shifts and window partitioning are relatively cheap and easily parallelizable operations. 
Additionally, it involves little to no changes to the self attention module, making implementation easier.
Swin became the state of the art across multiple vision tasks, and followed by Swin-V2~\cite{liu2022swin} to accommodate large scale pre-training.

Neighborhood Attention Transformer (NAT)~\cite{hassani2022neighborhood} was introduced later, with a simple sliding-window based attention, Neighborhood Attention (NA). 
Unlike Stand Alone Self Attention (SASA)~\cite{ramachandran2019stand}, which applies attention in the style of convolutions, NA localizes self attention to the nearest neighbors around each token, which allows it by definition to approach self attention and enjoy a fixed attention span. 
Such pixel-wise attention operations were assumed to be inefficient and challenging to parallelize~\cite{liu2021swin,ramachandran2019stand,vaswani2021scaling}, until the release of Neighborhood Attention Extension~\cite{hassani2022neighborhood}.
With this extension, NA can run even faster than Swin's SWSA in practice.
\looseness=-1 NAT was able to significantly outperform Swin on image classification, and achieved competitive performance on downstream tasks, while also scaling up to be even faster than Swin despite the slightly different architecture. 

Despite the efforts into hierarchical vision transformers with local attention, some of self attention’s most important properties, including \textit{global receptive field}, and the ability to \textit{model long-range inter-dependencies}, are weakened as a result of this localization. 

This leads to a simple question: \textbf{\textit{How does one maintain the tractability that local attention provides in hierarchical vision transformers, while avoiding its shortcomings?}}
In other words, the optimal scenario is maintaining the linear complexity, while preserving the global receptive field and the ability to model long-range inter-dependencies of self attention.
In this paper, we aim to answer this question and improve hierarchical transformers by extending a simple local attention mechanism, Neighborhood Attention, to Dilated Neighborhood Attention (DiNA): a flexible and powerful \textbf{sparse global attention}. 
Dilating neighborhoods in NA into larger sparse regions has multiple advantages: 1. it captures more global context, 2. allows the receptive field to grow exponentially, as opposed to linearly~\cite{yu2015multi}, and 3. comes \textit{at no additional computational cost}.
To demonstrate the effectiveness of DiNA, we propose Dilated Neighborhood Attention Transformer (DiNAT), which not only improves the existing NAT model in terms of downstream performance, it manages to outperform strong modern CNN baselines, such as ConvNeXt~\cite{liu2022convnet}, in downstream tasks with a noticeable margin.

\noindent Our main contributions can be summarized as follows:
\begin{itemize}
    \item Introducing DiNA, a simple and powerful sparse global attention pattern, which allows receptive field to grow exponentially and captures longer-range context without any additional computational burden. DiNA does so while maintaining the symmetry in neighborhoods introduced in NA. It can also adapt to larger resolutions without expanding to larger window sizes.
    \item Analyzing theoretical receptive field sizes in models based on convolutions, localized attention, and a DiNA-based model.
    \item Introducing DiNAT, a new hierarchical vision transformer made of both dilated and non-dilated variants of NA. DiNAT utilizes a \textit{gradual} dilation change through the model, which extends receptive fields more optimally and helps fine-to-coarse feature learning.
    \item Conducting extensive experiments on image classification, object detection, and segmentation with DiNAT, and finding that it exhibits a noticeable improvement in downstream tasks over both attention-based and convolutional baselines. Additionally, we investigate isotropic and hybrid attention variants, scaling experiments with ImageNet-22K pre-training, and the effects of different dilation values. We also achieve state of the art image segmentation performance with advanced segmentation frameworks.
    \item Extending \natten{}, NA's CUDA extension for PyTorch, by adding dilation support, and bfloat16 utilization, allowing the research in this direction to be extended to other tasks and applications.
\end{itemize}

While the initial experiments with DiNAT already exhibit significant improvements in downstream vision tasks, neither its performance nor applications stop here. 
NA's local attention and DiNA's sparse global attention complement each other: they can preserve locality, model longer-range inter-dependencies, expand the receptive field exponentially, and maintain a linear complexity.
Their restriction of self attention can potentially improve convergence by avoiding self attention’s possible redundant interactions, such as those with repetitive, background, or distracting tokens~\cite{liang2022evit,rao2021dynamicvit}.
Combinations of local attention and sparse global attention can potentially empower various vision tasks and beyond.
To support research in this direction, we open source our entire project, including our modified \natten{}, which can reduce runtime by orders of magnitude compared to naive implementations~\cite{hassani2022neighborhood}.
\section{Related Work}
\label{sec:related}
We briefly review dot product self attention (DPSA), the Transformer~\cite{vaswani2017attention}, and Vision Transformer~\cite{dosovitskiy2020image}.
We then move on to localized self attention modules such as SASA~\cite{ramachandran2019stand}, SWSA in Swin Transformer~\cite{liu2021swin}, and NA in Neighborhood Attention Transformer~\cite{hassani2022neighborhood}, and discuss their limitations, which are our motivation behind this work.
Finally, we discuss previous uses of sparse attention mechanisms in language processing~\cite{beltagy2020longformer,roy2021efficient} and vision~\cite{huang2019ccnet,child2019generating}.

\subsection{Self Attention}
Vaswani et al.~\cite{vaswani2017attention} define dot product attention as an operation between a query, and a set of key-value pairs. The dot product of the query and keys is scaled and sent through a softmax activation to produce attention weights. Said attention weights are then applied to the values:
\begin{equation}
    Attention ( Q, K, V ) = softmax \left( \frac{Q K^T}{\sqrt{d}} \right) V,
    \label{eq:attention}
\end{equation}
where $\sqrt{d}$ is the scaling parameter, and $d$ is the key dimension. Dot product self attention is simply a case of this operation where the queries, keys, and values are all linear projections of the same input.
Given an input $X \in \mathbb{R}^{n \times d}$, where $n$ is the number of tokens and $d$ is the embedding dimension, this operation has a complexity of $\mathcal{O}(n^2 d)$ and a space complexity of $\mathcal{O}(n^2)$ for the attention weights (space depends on implementation~\cite{dao2022flashattention}).
Vision Transformer (ViT)~\cite{dosovitskiy2020image} one of the earliest works applying a pure transformer encoder to vision, showed the power that a large scale self attention based model bears. 
Follow up works extended the study with minimal changes to training techniques~\cite{touvron2020training}, architectural changes~\cite{touvron2021going}, and applications to small data regimes~\cite{hassani2021escaping}.
Due to their quadratic time complexity, many works attempt to restrict the attention span in order to reduce compute, specifically when scaling to larger inputs, such as long documents in NLP~\cite{beltagy2020longformer}, and large resolutions in vision~\cite{liu2021swin}.
Restricting self attention can be done in different patterns, one of which is localization.

\subsection{Local Attention}

\paragraph{Stand-Alone Self Attention (SASA).} SASA~\cite{ramachandran2019stand} is one of the earliest local attention mechanisms that was specifically designed to be used in vision models, years before ViT~\cite{dosovitskiy2020image}.
It sets the key-value pair to sliding windows over the feature map, therefore localizing attention for each query (pixel) to a window centered around it.
Such an operation could easily replace convolutions in existing CNNs, such as ResNets, and theoretically even reduce computational complexity.
Despite the promise it showed, the authors found that the resulting model runs slow, due to the inefficient implementation of this module.
Works succeeding it therefore switched to alternative methods that could run more efficiently, such as blocked self attention in HaloNet~\cite{vaswani2021scaling}, and Window Self Attention in Swin~\cite{liu2021swin}.

\begin{figure}[t]
    \centering
    \includegraphics[width=0.47\textwidth]{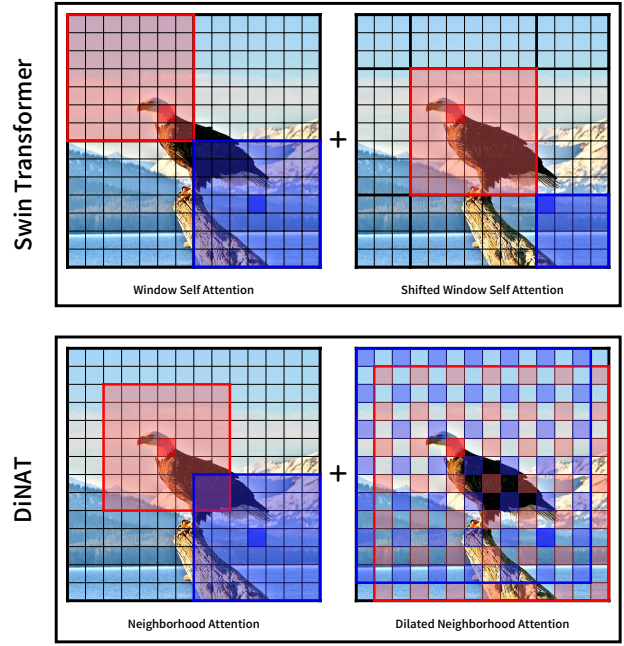}
    \caption{
    \textbf{Illustration of attention layers in Swin Transformer and DiNAT.} 
    Swin divides inputs into non-overlapping windows and applies self attention to each window separately, and applies a pixel shift every other layer. 
    Pixel-shifted layers mask attention weights between out-of-order regions, which restricts self attention to shifted subwindows.
    DiNAT applies Neighborhood Attention, a sliding window attention, and dilates it at every other layer.
    }
    \label{fig:dinatvsswin}
\end{figure}

\paragraph{Shifted Window Self Attention (SWSA).} Liu et al.~\cite{liu2021swin} proposed Window Self Attention (WSA) and its shifted variant SWSA, and used them in their hierarchical model for vision, Swin Transformer. 
They pointed out the inefficiency of sliding-window methods such as SASA as one of their motivations behind developing Window Self Attention.
The shifted variant (SWSA), as the name suggests, shifts pixels before the attention operation, and reverses the shift afterwards, to create a different window partitioning compared to the previous layer, which allows for out-of-window interactions that are crucial to a growing receptive field (see \cref{fig:dinatvsswin}).
Swin initially became the state of the art in object detection and semantic segmentation. It also inspired other works that extended it to different tasks beyond the ones explored in the paper, such as generation~\cite{zhang2022styleswin}, restoration~\cite{liang2021swinir}, masked image modeling~\cite{xie2022simmim}, video action recognition~\cite{liu2022video}, and more.
Additionally, the followup model, Swin-V2~\cite{liu2022swin}, became the new state of the art with their largest model.
It is noteworthy that Swin-V2 utilizes much larger window sizes to achieve such performance, which in turn increase time complexity and memory usage.

\paragraph{Neighborhood Attention (NA).} NA~\cite{hassani2022neighborhood} was proposed as a simple sliding-window attention, which localizes self attention for each pixel to its nearest neighbors. 
NA shares the same time and space complexity and number of parameters to those of Swin's WSA and SWSA, but instead operates in overlapping sliding windows, and therefore preserves translation equivariance.
While NA's sliding-window pattern is similar to SASA, its formulation of nearest neighbors makes it a direct restriction of self attention, and therefore NA, unlike SASA, approaches SA as its window size grows. 
A major challenge of sliding-window attention was the lack of efficient implementations, as no existing deep learning or CUDA libraries support such operations directly.
Therefore, NA was introduced along with \natten{}, an extension with efficient CPU and GPU kernels that allow NA to outperform modules such as WSA/SWSA in terms of both speed and memory usage.
The model Neighborhood Attention Transformer (NAT) is similar in its hierarchical design to Swin Transformer. The key differences, other than the attention modules, is that NAT utilizes overlapping convolutions in downsampling layers, as opposed to the patched ones used in Swin. 
\looseness=-1 As a result, to keep variants similar to Swin variants in terms of number of parameters and FLOPs, the models were made slightly deeper, with smaller inverted bottlenecks.
NAT achieves superior results in image classification compared to Swin, and performs competitively on downstream tasks.

While local attention based models are able to perform well across different vision tasks due to its preservation of locality and efficiency, they fall short of capturing global context like self attention, which is also crucial to vision.
Additionally, localized attention mechanisms utilize a smaller and slowly growing receptive field, similar to that of convolutions, compared to the full-sized receptive field in self attention.
Besides self attention, several works also explored global receptive fields in vision, including but not limited to Non-local Neural Networks~\cite{wang2018non}.
However, operations with unrestricted global receptive field usually suffer from high computational complexities compared to restricted ones, which can be local, or sparse.

\subsection{Sparse Attention}
Child et al.~\cite{child2019generating} proposed Sparse Transformers, which in addition to scaling to much deeper variants, utilized a sparse-kernel attention mechanism.
Through this, the model was able to train much more efficiently on longer sequences of data.
There have been other works in sparse attention, such as Longformer~\cite{beltagy2020longformer}, Routing Transformers~\cite{roy2021efficient}, and CCNet~\cite{huang2019ccnet}, all of which share a common feature: reducing the cost of self attention in cases where longer sequences of tokens are inevitable, but a global context is still necessary. Longformer~\cite{beltagy2020longformer} specifically investigates using a combination of 1-D sliding window attention with and without dilation, along with global attention for specific tokens. This results in a model that is able to process long documents while maintaining the global context.
CCNet~\cite{huang2019ccnet} uses axial attention to improve semantic segmentation heads by introducing global context without the quadratic cost of unrestricted self attention.
More recently, MaxViT~\cite{tu2022maxvit} explored a hybrid model, which uses a combination of MBConv, Window Attention~\cite{liu2021swin}, and sparse grid attention, obtaining high ImageNet accuracy. However, the resulting model yields higher complexity and lower throughput compared to Swin~\cite{liu2021swin}.

\begin{figure}[!t]
    \centering
    \includegraphics[page=1, width=0.475\textwidth]{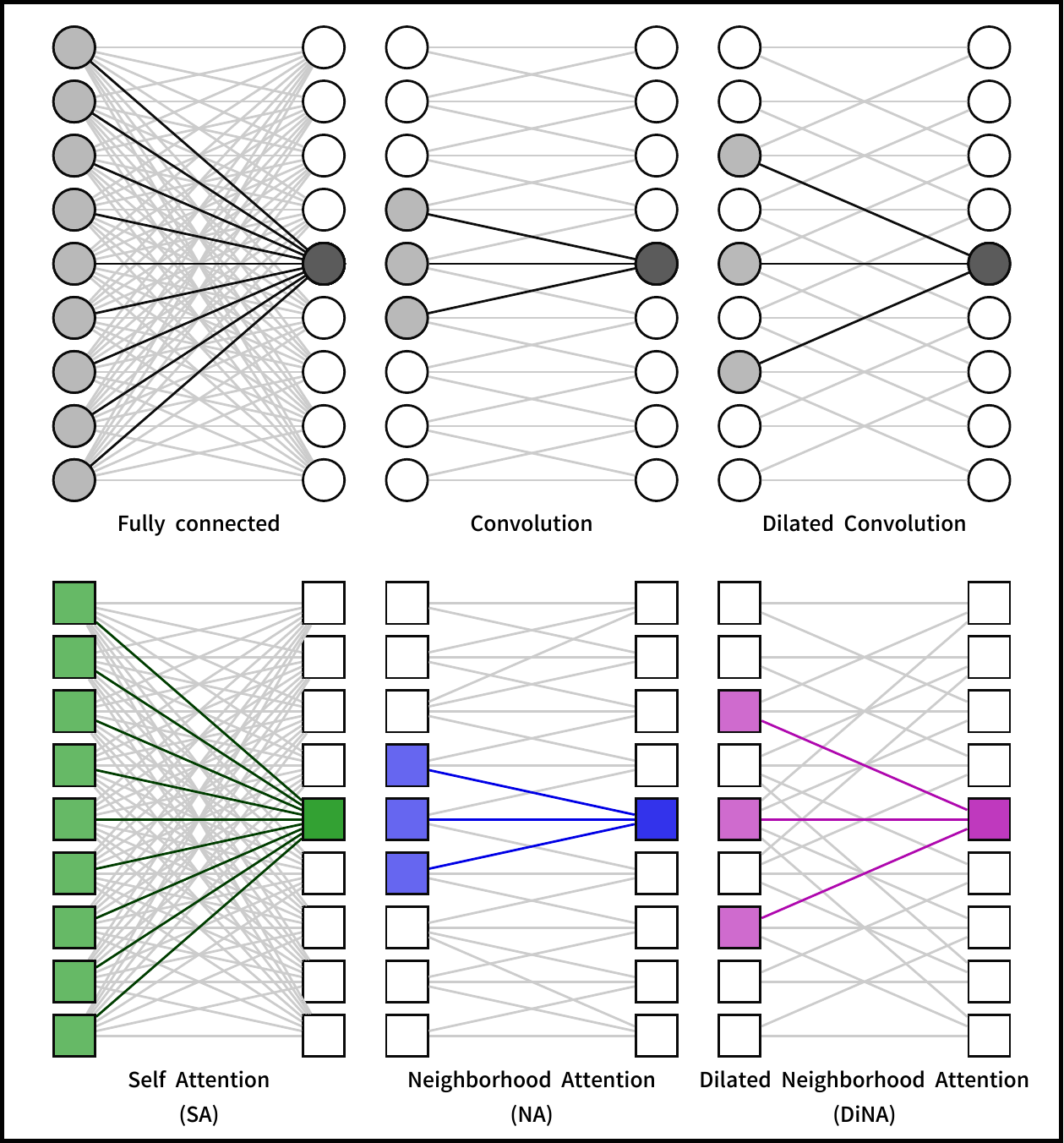}
    \caption{
    \textbf{A single-dimensional illustration of receptive fields in fully connected layers, convolutional layers, and different attention mechanisms.} 
    NA and DiNA restrict self attention through sliding windows, similar to how convolutions and dilated convolutions restrict fully connected layers. These restrictions reduce computational burden, introduce useful inductive biases, and in some cases increase flexibility w.r.t. varying input sizes.
    }
    \label{fig:connections}
\end{figure}

Even though such non-local and sparse restrictions of self attention have shown to be promising, they are not well-studied in the scope of hierarchical vision transformers.
To expand the local receptive fields, and re-introduce global context into hierarchical vision transformers, we introduce Dilated Neighborhood Attention (DiNA), an extension of NA that spans neighborhoods over longer ranges by increasing the step size, while maintaining the overall attention span.
DiNA can serve as a sparse and global operation and works most effectively when used in conjunction with NA as a local-only operation.
We present an illustration of receptive fields in \cref{fig:connections}, where we compare fully connected layers to convolutions and dilated convolutions, and similarly self attention, to NA and DiNA.
We provide empirical evidence for this claim with our hierarchical vision transformer, Dilated Neighborhood Attention Transformer (DiNAT).
\section{Method}
\label{sec:method}
In this section, we define DiNA as an extension to NA, analyze its effect on the receptive field, and move on to our model, DiNAT. 
We also provide brief details on implementation, and integration with the existing \natten{} package.

\begin{figure*}[!t]
    \centering
    \newcommand{\imgwidth}{0.195}
    \begin{subfigure}{\imgwidth\textwidth}
        \centering
        \includegraphics[page=5, width=1.0\textwidth]{figures/rf.pdf}
        \caption*{\textbf{ViT}.}
        \scriptsize {Complexity:} $\bigO \big( n^2d \big)$, \\{RF} $= n$.
    \end{subfigure}
    \begin{subfigure}{\imgwidth\textwidth}
        \centering
        \includegraphics[page=1, width=1.0\textwidth]{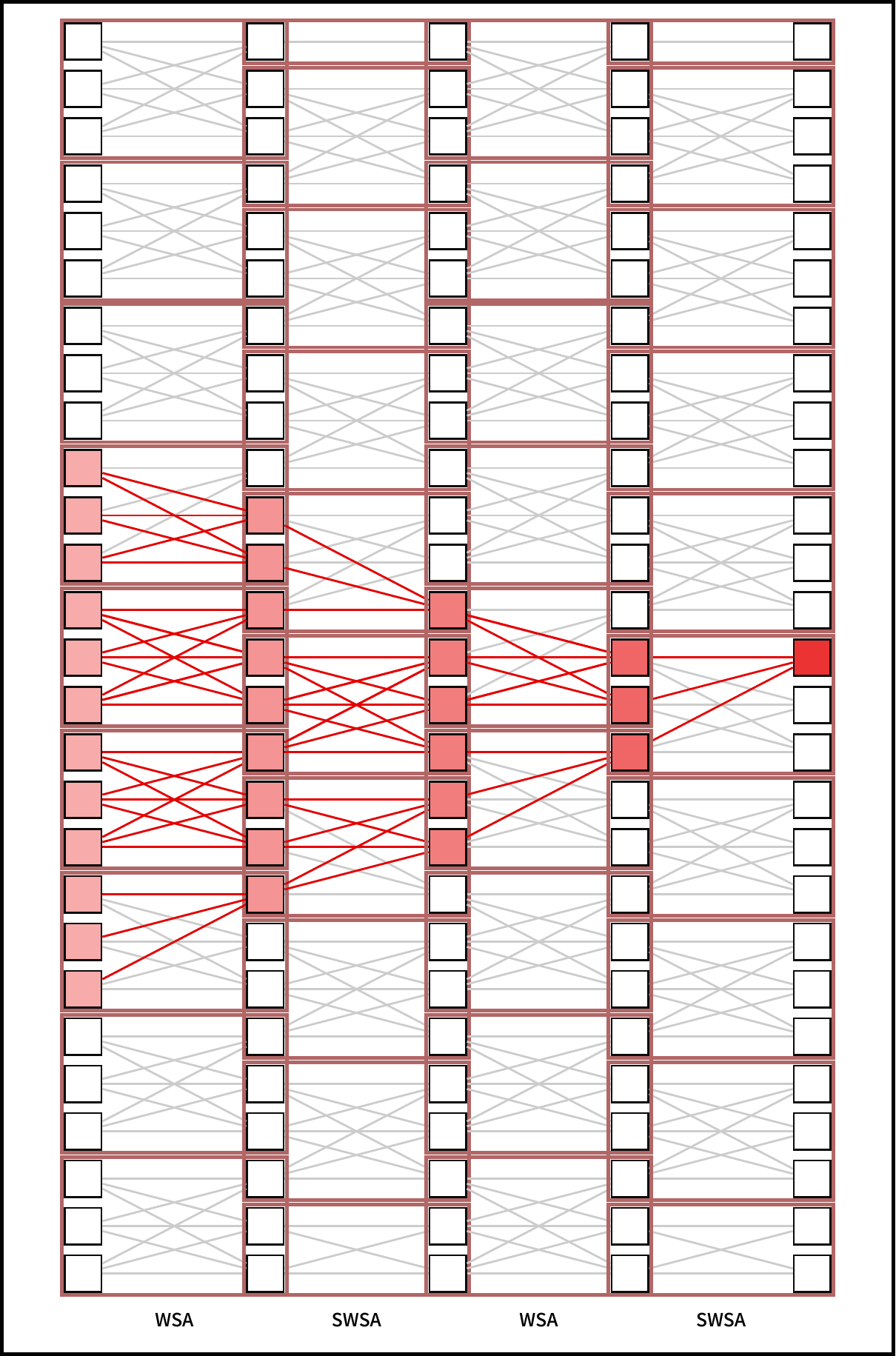}
        \caption*{\textbf{Swin}.}
        \scriptsize {Complexity:} $\bigO \big( ndk \big)$, \\{RF} $= \ell k$.
    \end{subfigure}
    \begin{subfigure}{\imgwidth\textwidth}
        \centering
        \includegraphics[page=3, width=1.0\textwidth]{figures/rf.pdf}
        \caption*{\textbf{ConvNeXt}.}
        \scriptsize {Complexity:} $\bigO \big( ndk \big)$, \\{RF} $= \ell (k - 1) + 1$.
    \end{subfigure}
    \begin{subfigure}{\imgwidth\textwidth}
        \centering
        \includegraphics[page=2, width=1.0\textwidth]{figures/rf.pdf}
        \caption*{\textbf{NAT}.}
        \scriptsize {Complexity:} $\bigO \big( ndk \big)$, \\{RF} $= \ell (k - 1) + 1$.
    \end{subfigure}
    \begin{subfigure}{\imgwidth\textwidth}
        \centering
        \includegraphics[page=4, width=1.0\textwidth]{figures/rf.pdf}
        \caption*{\textbf{DiNAT}.}
        \scriptsize {Complexity:} $\bigO \big( ndk \big)$, \\{RF} $\in [ \ell (k-1) + 1, k^\ell ]$.
    \end{subfigure}
    \caption{
    \textbf{Receptive fields in ViT, Swin, ConvNeXt, NAT, and our DiNAT.}
    We also provide the complexity of each method's primary operation. $n$ denotes the number of tokens, $d$ denotes the embedding dimension, and $k$ denotes kernel/window size.
    All receptive fields are bounded by input size, $n$.
    DiNAT's receptive field is flexible and ranges from linear, $\ell (k-1) + 1$, to exponential growth, $k^\ell$.
    }
    \label{fig:receptivefields}
\end{figure*}

\subsection{Dilated Neighborhood Attention}
For simplicity, we keep notations limited to single-dimensional NA and DiNA.
Given input $X \in \mathbb{R}^{n \times d}$, whose rows are $d$-dimensional token vectors,
and query and key linear projections of $X$, $Q$ and $K$, and relative positional biases between any two tokens $i$ and $j$, $B(i, j)$,
we define neighborhood attention weights for the $i$-th token with neighborhood size $k$, $\mathbf{A}_{i}^{k}$, as the matrix multiplication of the $i$-th token's query projection, and its $k$ nearest neighboring tokens' key projections:
\begin{equation}
    \mathbf{A}_{i}^{k} = \begin{bmatrix}Q_i K_{\rho_{1}{(i)}}^T + B_{(i,\rho_{1}{(i)})} \\Q_i K_{\rho_{2}{(i)}}^T + B_{(i,\rho_{2}{(i)})} \\ \vdots \\ Q_i K_{\rho_{k}{(i)}}^T + B_{(i,\rho_{k}{(i)})}\end{bmatrix},
    \label{eq:nattenq}
\end{equation}
where $\rho_{j}{(i)}$ denotes $i$'s $j$-th nearest neighbor.
We similarly define neighboring values, $\mathbf{V}_{i}^{k}$, as a matrix whose rows are the $i$-th token's $k$ nearest neighboring value projections:
\begin{equation}
    \mathbf{V}_{i}^{k} = \begin{bmatrix}V_{\rho_{1}{(i)}}^T & V_{\rho_{2}{(i)}}^T & \hdots & V_{\rho_{k}{(i)}}^T\end{bmatrix}^T,
    \label{eq:nattenv}
\end{equation}
where $V$ is a linear projection of $X$.
Neighborhood Attention output for the $i$-th token with neighborhood size $k$ is then defined as:
\begin{equation}
    \text{NA}_{k}{(i)} = softmax\left(\frac{\mathbf{A}_{i}^{k}}{\sqrt{d}}\right) \mathbf{V}_{i}^{k},
    \label{eq:natten}
\end{equation}
where $\sqrt{d}$ is the scaling parameter, and $d$ is the embedding dimension.
To extend this definition to DiNA, given a dilation value $\delta$, we simply define $\rho_{j}^{\delta}{(i)}$ as token $i$'s $j$-th nearest neighbor that satisfies: $j \bmod \delta = i \bmod \delta$.
We can then define \textbf{$\delta$-dilated} neighborhood attention weights for the $i$-th token with neighborhood size $k$, $\mathbf{A}_{i}^{(k, \delta)}$, as follows:
\begin{equation}
    \mathbf{A}_{i}^{(k, \delta)} = \begin{bmatrix}Q_i K_{\rho_{1}^{\delta}{(i)}}^T + B_{(i,\rho_{1}^{\delta}{(i)})} \\Q_i K_{\rho_{2}^{\delta}{(i)}}^T + B_{(i,\rho_{2}^{\delta}{(i)})} \\ \vdots \\ Q_i K_{\rho_{k}^{\delta}{(i)}}^T + B_{(i,\rho_{k}^{\delta}{(i)})}\end{bmatrix}.
    \label{eq:dinattenq}
\end{equation}
We similarly define $\delta$-dilated neighboring values for the $i$-th token with neighborhood size $k$, $\mathbf{V}_{i}^{(k, \delta)}$:
\begin{equation}
    \mathbf{V}_{i}^{(k, \delta)} = \begin{bmatrix}V_{\rho_{1}^{\delta}{(i)}}^T & V_{\rho_{2}^{\delta}{(i)}}^T & \hdots & V_{\rho_{k}^{\delta}{(i)}}^T\end{bmatrix}^T.
    \label{eq:dinattenv}
\end{equation}
DiNA output for the $i$-th token neighborhood size $k$ is then defined as:
\begin{equation}
    \text{DiNA}_{k}^{\delta}{(i)} = softmax\left(\frac{\mathbf{A}_{i}^{(k, \delta)}}{\sqrt{d_k}}\right) \mathbf{V}_{i}^{(k, \delta)}. 
    \label{eq:dinatten}
\end{equation}

\subsection{Choice of Dilation}

\setlength{\tabcolsep}{3pt}
\begin{table}[t]
    \centering
    \resizebox{0.475\textwidth}{!}{
    \begin{tabular}{lccl}
        \toprule
        \textbf{Layer structure}        & \textbf{Memory usage} & \textbf{FLOPs}        & \textbf{Receptive Field} \\
        \midrule
        \cb\cb \textbf{DWSConv-DWSConv} & $d^2 + d k$           & $nd^2 + n d k$        & $\ell (k - 1) + 1$ \\
        \midrule
        \wb\wb \textbf{WSA-WSA}         & $3d^2 + n k$          & $3nd^2 + 2 n d k$     & $k$ \\
        \wb\wb \textbf{WSA-SWSA}        & $3d^2 + n k$          & $3nd^2 + 2 n d k$     & $\ell k$ \\
        \midrule
        \nb\nb \textbf{NA-NA}           & $3d^2 + n k$          & $3nd^2 + 2 n d k$     & $\ell (k - 1) + 1$ \\
        \grayrow \nb\db \textbf{NA-DiNA}& $3d^2 + n k$          & $3nd^2 + 2 n d k$     & $\in [\ell (k-1) + 1, k^\ell]$ \\
        \midrule
        \vb\vb \textbf{SA-SA}           & $3d^2 + n^2$          & $3nd^2 + 2 n^2 d$     & $n$ \\
        \bottomrule
    \end{tabular}
    }
    \caption{
    \textbf{Memory usage (weights), FLOPs, and receptive field sizes in different models.} 
    Convolutions and NA expand receptive field linearly with model depth. 
    Window Self Attention alone would suffer from a fixed-value receptive field, but the pixel shift in SWSA expands the receptive field linearly.
    NA and DiNA together can expand receptive fields \textit{exponentially}.
    Self attention has the maximum receptive field, which comes at the expense of a quadratic computational cost.
    Note that the denoted receptive fields have an upper bound of $n$.
    }
    \label{tab:receptive}
\end{table}
DiNA introduces a key new architectural hyperparameter: per layer dilation values. 
We define the upper bound for dilation value to be $\lfloor \frac{n}{k} \rfloor$, where $n$ is the number of tokens, and $k$ is kernel/neighborhood size.
This is simply to ensure exactly $k$ dilated neighbors exist for each token.
The lower bound is always 1, which would be equivalent to vanilla NA.
Therefore, dilation value in each layer of the model will be an input-dependent hyperparameter, which can take any integer $\delta \in [ 1,  \lfloor \frac{n}{k} \rfloor ]$.
Because dilation values are changeable, they provide a flexible receptive field (discussed in \cref{subsec:receptive}).
It is not feasible to try out all possible combinations, therefore we explored a limited number of choices, which are discussed in \cref{subsec:miscexps}.

\begin{figure*}[t]
    \centering
    \includegraphics[width=1.0\textwidth]{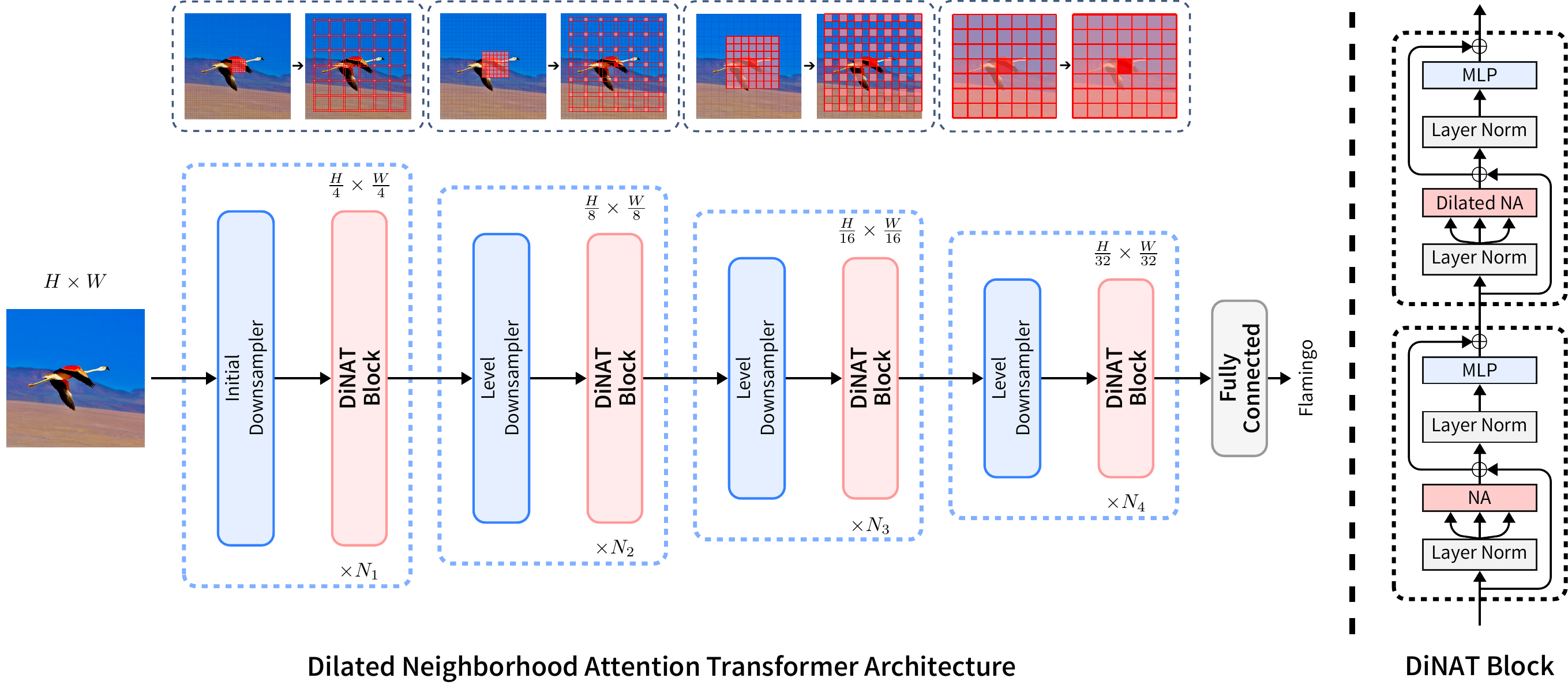}
    \caption{
    \textbf{An illustration of DiNAT's architecture.} 
    It downsamples inputs to a quarter of their original spatial resolution initially, and sends them through 4 levels of DiNA Transformer encoders.
    Feature maps are downsampled to half their spatial size and doubled in channels between levels.
    DiNAT layers are similar to most Transformers: Attention followed by an MLP with normalization and skip connections in between.
    It also switches between local NA and sparse global DiNA at every other layer (right).
    }
    \label{fig:model_overview}
\end{figure*}

\subsection{Receptive Fields}
\label{subsec:receptive}
We analyze DiNA’s receptive field, as it is important to understanding the power of DiNA, especially in comparison to other models.
We present a comparison of receptive field sizes in different attention patterns in \cref{tab:receptive}, along with FLOPs and memory usage.
We also include depth-wise separable convolution (DWSConv), the key component in ConvNeXt~\cite{liu2022convnet}, for completeness.

We calculate receptive field size with respect to the number of layers, $\ell$, kernel size $k$, and number of tokens $n$.
Both convolutions and NA start out with a receptive field of size $k$, and expand by $k - 1$ per layer (center pixel remains fixed).
Swin Transformer's Window Self Attention~\cite{liu2021swin} on its own maintains a constant receptive field size, as the window partitioning prevents cross-window interactions, hence preventing receptive field expansion.
Pixel shifted WSA resolves this issue, and expands receptive fields by exactly one window per layer, which is an expansion of $k$ per layer. 

It is worth noting that while Swin enjoys a slightly larger receptive field compared to NAT and ConvNeXt thanks to its special shifted window design, it breaks an important property: \textit{symmetry}. 
Since Swin's feature maps are partitioned into non-overlapping windows, pixels within the same window only attend to each other, regardless of their position (whether at center or corner), leading to some pixels seeing asymmetric context around them.

Unlike the fixed receptive field growth in NAT, Swin, and ConvNeXt, DiNA's receptive field is flexible and changes with dilation. 
It can range anywhere from NAT's original $\ell (k - 1) + 1$ (all dilation values set to 1), to and exponentially growing receptive field of $k^\ell$ (gradual dilation increase), which is one of the main reasons behind its power.
Regardless of dilation, the first layer always yields a receptive field of size $k$.
Given large enough dilation values, the preceding DiNA layer will yield a $k$-sized receptive field for each of the $k$ in the DiNA layer, yielding a receptive field of size $k^2$.
As a result, DiNA and NA combinations with optimal dilation values can potentially increase receptive field \textit{exponentially} to $k^\ell$.
This comes with no surprise, as dilated convolutions have also been known for having an exponentially-growing receptive field size when using exponentially growing dilation values~\cite{yu2015multi}.
An illustration of the increased receptive field size is also presented in \cref{fig:receptivefields}.

\subsection{DiNAT}
\setlength{\tabcolsep}{6pt}
\begin{table}[t]
    \centering
    \resizebox{0.475\textwidth}{!}{
    \begin{tabular}{lccccc}
        \toprule
        \textbf{Variant}                        & \textbf{Layers}   & \textbf{Dim \texttimes{}}   & \textbf{MLP}                & \textbf{\# of}     & \textbf{FLOPs}\\
                                                & \textbf{per level}& \textbf{Heads}              & \textbf{ratio}              & \textbf{Params}    &               \\
        \midrule
        \db\textbf{DiNAT-Mini}           & 3, 4,  6, 5       & 32 \texttimes{} 2                 & 3                     &  20 M                 &  2.7 G \\
        \db\textbf{DiNAT-Tiny}           & 3, 4, 18, 5       & 32 \texttimes{} 2                 & 3                     &  28 M                 &  4.3 G \\
        \db\textbf{DiNAT-Small}          & 3, 4, 18, 5       & 32 \texttimes{} 3                 & 2                     &  51 M                 &  7.8 G \\
        \db\textbf{DiNAT-Base}           & 3, 4, 18, 5       & 32 \texttimes{} 4                 & 2                     &  90 M                 & 13.7 G \\
        \db\textbf{DiNAT-Large}          & 3, 4, 18, 5       & 32 \texttimes{} 6                 & 2                     & 200 M                 & 30.6 G \\
        \bottomrule
    \end{tabular}
    }
    \caption{\textbf{DiNAT variants.} 
    In terms of architecture, DiNAT is identical to NAT, which follows Swin closely in overall design.
    Channels (heads and dim) double after every level. Kernel size is 7\sq in all variants.
    }
    \label{tab:variants}
\end{table}

For a fair evaluation of DiNA's performance, we design DiNAT to be identical to the original NAT model in terms of architecture and configuration. 
It uses two 3\texttimes{}3 convolutional layers with 2\texttimes{}2 strides initially, resulting in feature maps that are a quarter of the input resolution.
It also uses a single 3\texttimes{}3 convolution with 2\texttimes{}2 strides to downsample between levels, which cut spatial resolution in half and double channels.
Details are presented in \cref{tab:variants}.
The key difference in DiNAT is that every other layer uses DiNA instead of NA.
Dilation values for DiNA layers are set based on the task and input resolution. For ImageNet-1k at 224\sq{} resolution, we set dilation values to 8, 4, 2, and 1 in levels one through four respectively. 
In downstream tasks, because of their larger resolution, we increase dilation values to beyond that.
All dilation values and other relevant architecture details are presented in \cref{apptab:dinatsettings}.

\subsection{Implementation}
\label{subsec:implementation}
We implemented DiNA on top of the existing Neighborhood Attention Extension (\natten{}), allowing ease of use and identical memory usage to NA.
The latest public version of the extension includes a more efficient ``tiled'' implementation of Neighborhood Attention, which is what allows it to compete with methods such as Swin in terms of speed. 
By adding a dilation element to all the existing CUDA kernels, and re-implementing the ``tiled'' kernel to support dilated memory format, we managed to implement DiNA without affecting the speed of the existing NA kernels. 
However, it should be noted that DiNA's throughput will depend on dilation value, and is expected to be slightly slower than NA in practice. 
This is simply due to the break in memory access pattern, which would affect throughput overall (see~\cref{appfig:cudatime}). 
We also note that these implementations are still fairly naive and don't fully utilize newer architecture standards in CUDA, such as Tensor Cores, and are therefore only working as a proof of concept.
Despite this limitation, models using NA and DiNA can achieve competitive throughput levels compared to other methods that mostly utilize convolutions, linear projections, and self attention, all of which run through NVIDIA libraries that fully utilize the aforementioned standards.
More information on implementation is provided in \cref{appsec:implementation}.
\section{Experiments}
We conducted extensive experiments to study the effects of our proposed DiNAT model over existing baselines.
Similar to existing methods, we pre-train models on image classification (ImageNet-1K and ImageNet-22K~\cite{deng2009imagenet}), and then transfer the learned weights to downstream vision tasks.
We compare DiNAT to the original NAT model~\cite{hassani2022neighborhood}, Swin~\cite{liu2021swin}, and ConvNeXt~\cite{liu2022convnet}.
We also pair our model with Mask2Former~\cite{cheng2022masked} and perform instance, semantic, and panoptic segmentation experiments.

\subsection{Image Classification}

\setlength{\tabcolsep}{8pt}
\begin{table}[t]
    \centering
    \resizebox{0.475\textwidth}{!}{
    \begin{tabular}{lccccc}
        \toprule
        \textbf{Model}& \textbf{\# of}     & \textbf{FLOPs} & \textbf{Thru.} & \textbf{Memory} & \textbf{Top-1}\\
                      & \textbf{Params}    &                & (imgs/sec)     & (GB)            & (\%) \\
        \midrule
        \multicolumn{6}{c}{\textit{ImageNet-1K trained models}}\\
        \midrule
        \nb\textbf{NAT-M}                           &  20 M &   2.7 G & 2132 &  2.4 & 81.8\\
        \ours \textbf{DiNAT-M}                      &  20 M &   2.7 G & 2080 &  2.4 & 81.8\\
        \midrule
        \wb\textbf{Swin-T}                          &  28 M &   4.5 G & 1724 &  4.8 & 81.3 \\
        \cb\textbf{ConvNeXt-T}                      &  28 M &   4.5 G & 2491 &  3.4 & 82.1 \\
        \nb\textbf{NAT-T}                           &  28 M &   4.3 G & 1537 &  2.5 & \textbf{83.2} \\
        \ours \textbf{DiNAT-T}                      &  28 M &   4.3 G & 1500 &  2.5 & 82.7 \\
        \midrule
        \wb\textbf{Swin-S}                          &  50 M &   8.7 G & 1056 &  5.0 & 83.0 \\
        \cb\textbf{ConvNeXt-S}                      &  50 M &   8.7 G & 1549 &  3.5 & 83.1 \\
        \nb\textbf{NAT-S}                           &  51 M &   7.8 G & 1049 &  3.7 & 83.7 \\
        \ours \textbf{DiNAT-S}                      &  51 M &   7.8 G & 1058 &  3.7 & \textbf{83.8} \\
        \midrule
        \wb\textbf{Swin-B}                          &  88 M &  15.4 G &  774 &  6.7 & 83.5 \\
        \cb\textbf{ConvNeXt-B}                      &  89 M &  15.4 G & 1107 &  4.8 & 83.8 \\
        \nb\textbf{NAT-B}                           &  90 M &  13.7 G &  781 &  5.0 & 84.3 \\
        \ours \textbf{DiNAT-B}                      &  90 M &  13.7 G &  764 &  5.0 & \textbf{84.4} \\
        \midrule
        \multicolumn{6}{c}{\textit{ImageNet-22K pre-trained models}}\\
        \midrule
        \wb\textbf{Swin-L}                          & 197 M &  34.5 G &  478 & 10.4 & 86.3 \\
        \cb\textbf{ConvNeXt-L}                      & 198 M &  34.4 G &  643 &  7.5 & \textbf{86.6} \\
        \ours \textbf{DiNAT-L}                      & 200 M &  30.6 G &  474 &  7.8 & \textbf{86.6} \\
        \bottomrule
    \end{tabular}
    }
    \caption{
    \textbf{ImageNet-1K image classification performance at 224\sq{} resolution.} 
    \imagenetnotes{}
    }
    \label{tab:imagenet_comparison}
\end{table}

\setlength{\tabcolsep}{3pt}
\begin{table}[t]
    \centering
    \resizebox{0.475\textwidth}{!}{
    \begin{tabular}{lcccccc}
        \toprule
        \textbf{Model} & \textbf{Win.} & \textbf{\# of}     & \textbf{FLOPs} & \textbf{Thru.} & \textbf{Memory} & \textbf{Top-1}\\
                       & \textbf{Size} & \textbf{Params}    &                & (imgs/sec)     & (GB)            & (\%) \\
        \midrule
        \wb\textbf{Swin-L}                          & 12\sq & 197 M & 104.0 G &  169 & 32.7 & 87.3 \\
        \cb\textbf{ConvNeXt-L}                      &  7\sq & 198 M & 101.1 G &  221 & 19.2 & \textbf{87.5} \\
        \ours \textbf{DiNAT-L}                  &  7\sq & 200 M &  89.7 G &  161 & 20.1 & 87.4 \\
        \ours \textbf{DiNAT-L}                  & 11\sq & 200 M &  92.4 G &  110 & 26.9 & \textbf{87.5} \\
        \bottomrule
    \end{tabular}
    }
    \caption{
    \textbf{ImageNet-1K image classification performance at 384\sq{} resolution.} 
    \imagenetnotes{}
    }
    \label{tab:imagenet_comparison_22k}
\end{table}

We used the community standard for ImageNet training in PyTorch, \verb|timm|~\cite{rw2019timm} (Apache License v2), which now serves as the community standard for ImageNet training in PyTorch~\cite{paszke2019pytorch}, to train our model on ImageNet-1k~\cite{deng2009imagenet}. 
We use the same training configurations, regularization techniques, and augmentations (CutMix \cite{yun2019cutmix}, Mixup \cite{zhang2017mixup}, RandAugment \cite{cubuk2020randaugment}, and Random Erasing \cite{zhong2020random}) and training techniques used in NAT~\cite{hassani2022neighborhood} and Swin~\cite{liu2021swin}.
Models trained on ImageNet-1K directly are trained for 300 epochs with a batch size of 1024, and use an iteration-wise cosine learning rate schedule and a 20 epoch warmup, with a base learning rate of 1e-3, and weight decay rate of 0.05, cooled down for an additional 10 epochs. 
Larger variants are pre-trained on ImageNet-22K~\cite{deng2009imagenet} for 90 epochs with a batch size of 4096, but use a linear learning rate schedule and a 5 epoch warmup, with a base learning rate of 1e-3, and weight decay rate of 0.01, again following Swin~\cite{liu2021swin}.
We fine-tune models pre-trained on ImageNet-22K to ImageNet-1K for 30 epochs, with a batch size of 512, and a linear learning rate schedule with no warmup, and a base learning rate of 5e-5, and weight decay rate of 1e-4.
Final ImageNet-1K validation set accuracy levels, along with number of learnable parameters, FLOPs, throughput, and memory usage are provided in \cref{tab:imagenet_comparison,tab:imagenet_comparison_22k}.
The reason for providing both FLOPs and throughput is to point out the necessity in distinguishing theoretical computational requirements, versus efficiency in practice with each method's available implementation.
This is especially important in this case because NA and DiNA are based on from scratch implementations of the algorithms (\natten{}), and are not as well-optimized as ConvNeXt or Swin, which mostly run on native NVIDIA libraries designed for optimal throughput.

\paragraph{ImageNet-1K. } DiNAT doesn't show improvement over NAT in smaller variants. 
Improvement over NAT-Mini is less than 0.1\%, and we found that while the Tiny variant converges faster than NAT-Tiny at first, it converges to a lower accuracy of 82.7\%.
We noticed that despite this, DiNAT consistently outperforms NAT across all four variants on downstream tasks.
DiNAT shows a slight improvement of at least 0.1\% over NAT on Small and Base variants.

\paragraph{ImageNet-22K. } We pre-trained our Large variant on ImageNet-22K, and fine-tuned it to ImageNet-1K at both 224\sq{} and 384\sq{} resolutions.
We found that our large variant can successfully outperform Swin-Large and match ConvNeXt-Large's accuracy at 224\sq{} resolution.
At 384\sq{}, our large variant exceeds its Swin counterpart's reported accuracy without increasing its kernel size from 7\sq{} to 12\sq{}. 
Upon increasing the large variant's kernel size to 11\sq{} and interpolating positional biases (similar to Swin), we see that our large variant matches ConvNeXt-Large's accuracy as well. 
We note that NA/DiNA are in theory limited to odd-sized kernels, which is the reason behind picking 11\sq{} instead of 12\sq{}.

\paragraph{Isotropic variants.} To further compare NA/DiNA to plain self attention, we also explore \textit{isotropic} variants of NAT and DiNAT, similar to isotropic ConvNeXt~\cite{liu2022convnet} variants.
These models simply follow ViT in design: a single Transformer encoder operating on feature maps with a fixed spatial size (14\sq), preceded by a single patch-and-embedding layer; they are not hierarchical transformers.
To maintain fairness in comparison to self attention, we trained ViT models with relative positional biases (ViT\pls) to ensure the models are only different in attention patterns.
Note that ViT variants with relative positional biases have previously been explored in timm~\cite{rw2019timm}, but we run our own to ensure similar training settings.
We present a comparison of these models and their performance on ImageNet-1k in \cref{tab:isotropic}.
\setlength{\tabcolsep}{3pt}
\begin{table}[t]
    \centering
    \resizebox{0.475\textwidth}{!}{
    \begin{tabular}{lcccccc}
        \toprule
        \textbf{Model} & \textbf{\# of}     & \textbf{FLOPs} & \textbf{Thru.} & \textbf{Memory} & \textbf{Top-1}\\
                       & \textbf{Params}    &                & (imgs/sec)     & (GB)            & (\%) \\
        \midrule
        \cb\textbf{ConvNeXt-S}\iso                  &  22 M &   4.3 G & 4327 &  1.2 & 79.7 \\
        \nb\textbf{NAT-S}\iso                       &  22 M &   4.3 G & 3255 &  1.3 & 80.0 \\
        \ours \textbf{DiNAT-S}\iso                  &  22 M &   4.3 G & 3160 &  1.3 & 80.8 \\
        \vb\textbf{ViT\pls-S}                       &  22 M &   4.6 G & 3086 &  1.9 & \textbf{81.2} \\
        \midrule
        \cb\textbf{ConvNeXt-B}\iso                  &  87 M &  16.9 G & 1661 &  2.4 & 82.0 \\
        \nb\textbf{NAT-B}\iso                       &  86 M &  16.9 G & 1350 &  2.7 & 81.6 \\
        \ours \textbf{DiNAT-B}\iso                  &  86 M &  16.9 G & 1316 &  2.7 & 82.1 \\
        \vb\textbf{ViT\pls-B}                       &  86 M &  17.5 G & 1284 &  3.7 & \textbf{82.5} \\
        \bottomrule
    \end{tabular}
    }
    \caption{
    \textbf{ImageNet-1K Top-1 validation accuracy comparison of ConvNeXt, NAT, and DiNAT's isotropic variants to ViT.} 
    \isonotes{}
    \imagenetnotes{}
    }
    \label{tab:isotropic}
\end{table}
We find that isotropic variants of both NAT and DiNAT exhibit only minor throughput improvements over ViT\pls, which can again be attributed to the lack of fully optimized implementations.
Note that these variants reduce FLOPs to almost the same number as isotropic ConvNeXt variants.
They also reduce memory usage compared to ViT\pls noticeably.
As for performance, we observe that isotropic NAT variants result in a drop in performance compared to ViT\pls, which is to be expected since NAT has half the attention span as ViT\pls.
However, we find that isotropic DiNAT variants significantly improve upon NAT's isotropic variants, without increasing kernel size.
This further supports our claim that a combination of NA and DiNA is more effective at producing an alternative to self attention than simply using NA throughout the model.
\setlength{\tabcolsep}{8pt}
\begin{table}[t]
    \centering
    \resizebox{0.475\textwidth}{!}{
    \begin{tabular}{lccc}
        \toprule
        \textbf{Model} & \textbf{Layer structure}     & \textbf{FLOPs} & \textbf{Top-1} (\%)\\
        \midrule
        \nb\nb\textbf{NAT-S}\iso                    & NA-NA        &   4.32 G   & 80.0 \\
        \db\db                                      & DiNA-DiNA    &   4.32 G   & 77.9 \\
        \db\nb                                      & DiNA-NA      &   4.32 G   & 80.6 \\
        \grayrow\nb\db \textbf{DiNAT-S}\iso         & NA-DiNA      &   4.32 G   & \textbf{80.8} \\
        \midrule
        \vb\nb                                      & SA-NA        &   4.45 G   & 81.0 \\
        \vb\db                                      & SA-DiNA      &   4.45 G   & 81.1 \\
        \nb\vb                                      & NA-SA        &   4.45 G   & \textbf{81.2} \\
        \db\vb                                      & DiNA-SA      &   4.45 G   & 80.9 \\
        \midrule
        \vb\vb\textbf{ViT\pls-S}                    & SA-SA        &   4.58 G   & \textbf{81.2} \\
        \bottomrule
    \end{tabular}
    }
    \caption{
    \textbf{Comparison of different layer structures in the isotropic variant.} 
    We compare different attention mechanisms in detail by creating hybrid models with both SA and NA/DiNA.
    }
    \label{tab:isotropic_ablation}
\end{table}
To further study the effects of different attention mechanisms, and investigate whether or not a model fully based on self-attention always yields the best result, we experiment with hybrid isotropic models utilizing both NA/DiNA layers as well as self attention. We present those results in \cref{tab:isotropic_ablation}.
We found that a small-scale (22M parameter) model with only half the layers performing self attention and the other half neighborhood attention can reach a similar accuracy as a similar model with all 12 layers utilizing self attention. 
We also found that changing the order of different attention layers can result in an approximately 0.2\% change in accuracy.

\subsection{Object Detection and Instance Segmentation}
\label{sec:detection}

\setlength{\tabcolsep}{3pt}
\begin{table}[t]
    \centering
    \resizebox{0.475\textwidth}{!}{
    \begin{tabular}{lccc|ccc|ccc}
        \toprule
        \textbf{Backbone} & \textbf{\# of} & \textbf{FLOPs} & \textbf{Thru.} & \textbf{AP\textsuperscript{b}} & \textbf{AP\textsuperscript{b}\textsubscript{50}} & \textbf{AP\textsuperscript{b}\textsubscript{75}} & \textbf{AP\textsuperscript{m}} & \textbf{AP\textsuperscript{m}\textsubscript{50}} & \textbf{AP\textsuperscript{m}\textsubscript{75}} \\ 
        & \textbf{Params} &&(FPS)\\
        \midrule
        \multicolumn{10}{c}{\textit{Mask R-CNN - 3x schedule}} \\
        \midrule
        \nb\textbf{NAT-M}                                       &  40 M &  225 G & 54.1 & 46.5 & 68.1 & 51.3 & 41.7 & 65.2 & 44.7 \\
        \ours \textbf{DiNAT-M}                                  &  40 M &  225 G & 53.8 & \textbf{47.2} & \textbf{69.1} & \textbf{51.9} & \textbf{42.5} & \textbf{66.0} & \textbf{45.9} \\
        \midrule
        \wb\textbf{Swin-T}                                      &  48 M &  267 G & 45.1 & 46.0 & 68.1 & 50.3 & 41.6 & 65.1 & 44.9 \\
        \cb\textbf{ConvNeXt-T}                                  &  48 M &  262 G & 52.0 & 46.2 & 67.0 & 50.8 & 41.7 & 65.0 & 44.9 \\
        \nb\textbf{NAT-T}                                       &  48 M &  258 G & 44.5 & 47.7 & 69.0 & 52.6 & 42.6 & 66.1 & 45.9 \\
        \ours \textbf{DiNAT-T}                                  &  48 M &  258 G & 43.3 & \textbf{48.6} & \textbf{70.2} & \textbf{53.4} & \textbf{43.5} & \textbf{67.3} & \textbf{46.8} \\
        \midrule
        \wb\textbf{Swin-S}                                      &  69 M &  359 G & 31.7 & 48.5 & 70.2 & 53.5 & 43.3 & 67.3 & 46.6 \\
        \nb\textbf{NAT-S}                                       &  70 M &  330 G & 34.8 & 48.4 & 69.8 & 53.2 & 43.2 & 66.9 & 46.5 \\
        \ours \textbf{DiNAT-S}                                  &  70 M &  330 G & 35.3 & \textbf{49.3} & \textbf{70.8} & \textbf{54.2} & \textbf{44.0} & \textbf{68.0} & \textbf{47.4} \\
        \midrule
        \multicolumn{10}{c}{\textit{Cascade Mask R-CNN - 3x schedule}} \\
        \midrule
        \nb\textbf{NAT-M}                                       &  77 M &  704 G & 27.8 & 50.3 & 68.9 & 54.9 & 43.6 & 66.4 & 47.2 \\
        \ours \textbf{DiNAT-M}                                  &  77 M &  704 G & 27.6 & \textbf{51.2} & \textbf{69.8} & \textbf{55.7} & \textbf{44.4} & \textbf{67.3} & \textbf{47.8} \\
        \midrule
        \wb\textbf{Swin-T}                                      &  86 M &  745 G & 25.1 & 50.4 & 69.2 & 54.7 & 43.7 & 66.6 & 47.3 \\
        \cb\textbf{ConvNeXt-T}                                  &  86 M &  741 G & 27.3 & 50.4 & 69.1 & 54.8 & 43.7 & 66.5 & 47.3 \\
        \nb\textbf{NAT-T}                                       &  85 M &  737 G & 24.9 & 51.4 & 70.0 & 55.9 & 44.5 & 67.6 & 47.9 \\
        \ours \textbf{DiNAT-T}                                  &  85 M &  737 G & 25.0 & \textbf{52.2} & \textbf{71.0} & \textbf{56.8} & \textbf{45.1} & \textbf{68.3} & \textbf{48.8} \\
        \midrule
        \wb\textbf{Swin-S}                                      & 107 M &  838 G & 20.3 & 51.8 & 70.4 & 56.3 & 44.7 & 67.9 & 48.5 \\
        \cb\textbf{ConvNeXt-S}                                  & 108 M &  827 G & 23.0 & 51.9 & 70.8 & 56.5 & 45.0 & 68.4 & 49.1 \\
        \nb\textbf{NAT-S}                                       & 108 M &  809 G & 21.7 & 52.0 & 70.4 & 56.3 & 44.9 & 68.1 & 48.6 \\
        \ours \textbf{DiNAT-S}                                  & 108 M &  809 G & 21.8 & \textbf{52.9} & \textbf{71.8} & \textbf{57.6} & \textbf{45.8} & \textbf{69.3} & \textbf{49.9} \\
        \midrule
        \wb\textbf{Swin-B}                                      & 145 M &  982 G & 17.3 & 51.9 & 70.9 & 56.5 & 45.0 & 68.4 & 48.7 \\
        \cb\textbf{ConvNeXt-B}                                  & 146 M &  964 G & 19.5 & 52.7 & 71.3 & 57.2 & 45.6 & 68.9 & 49.5 \\
        \nb\textbf{NAT-B}                                       & 147 M &  931 G & 18.6 & 52.3 & 70.9 & 56.9 & 45.1 & 68.3 & 49.1 \\
        \ours \textbf{DiNAT-B}                                  & 147 M &  931 G & 18.5 & \textbf{53.4} & \textbf{72.1} & \textbf{58.2} & \textbf{46.2} & \textbf{69.7} & \textbf{50.2} \\
        \midrule
        \wb\textbf{Swin-L\strr\dgg}                             & 253 M & 1393 G & 12.9 & 53.7 & 72.2 & 58.7 & 46.4 & 69.9 & 50.7 \\
        \cb\textbf{ConvNeXt-L\dgg}                              & 253 M & 1354 G & 14.8 & 54.8 & 73.8 & 59.8 & 47.6 & 71.3 & 51.7 \\
        \ours \textbf{DiNAT-L\dgg}                              & 258 M & 1276 G & 14.0 & \textbf{55.3} & \textbf{74.3} & \textbf{60.2} & \textbf{47.8} & \textbf{71.8} & \textbf{52.0} \\
        \bottomrule
    \end{tabular}
    }
    \caption{
    \textbf{COCO object detection and instance segmentation performance.} 
    \dgg indicates that the model was pre-trained on ImageNet-22K. 
    \strr Swin-L was not reported with Cascade Mask R-CNN, therefore we trained it with their official checkpoint. 
    \downstreamnotes{}
    }
    \label{tab:objectdetection}
\end{table}

To explore DiNAT's effectiveness in object detection and instance segmentation, we used its pre-trained weights as backbones for Mask R-CNN~\cite{he2017mask} and Cascade Mask R-CNN~\cite{cai2018cascade}, and trained those models on MS-COCO~\cite{lin2014microsoft}.
We followed NAT~\cite{hassani2022neighborhood} and Swin~\cite{liu2021swin}'s training settings in \verb|mmdetection| \cite{chen2019mmdetection} (Apache License v2), and trained with the same accelerated $3\times$ LR schedule. The results are presented in \cref{tab:objectdetection}. 
We observe that DiNAT consistently shows noticeable improvement over NAT, with little-to-no drop in throughput. 
There are even instances where DiNAT even surpasses NAT's throughput, but within the margin of error.
Additionally, we observe that this improvement over NAT pushes DiNAT ahead of ConvNeXt~\cite{liu2022convnet}.
At scale, we see DiNAT continues to outperform both Swin and ConvNeXt with ImageNet-22K pre-training.

\subsection{Semantic Segmentation}
\label{sec:segmentation}

\setlength{\tabcolsep}{5pt}
\begin{table}[t]
    \centering
    \resizebox{0.475\textwidth}{!}{
    \begin{tabular}{lcccccc}
        \toprule
        \textbf{Backbone} & \textbf{Res.} & \textbf{\# of}      & \textbf{FLOPs} & \textbf{Thru.} & \textbf{mIoU} \\ 
                          &               & \textbf{Params}     &                & (FPS)          &                     \\
        \midrule
        \nb\textbf{NAT-M}                       & 2048 \texttimes{} 512 &  50 M &  900 G & 24.5 & 46.4 \\
        \ours \textbf{DiNAT-M}                  & 2048 \texttimes{} 512 &  50 M &  900 G & 24.2 & \textbf{47.2} \\
        \midrule
        \wb\textbf{Swin-T}                      & 2048 \texttimes{} 512 &  60 M &  946 G & 21.3 & 45.8 \\
        \cb\textbf{ConvNeXt-T}                  & 2048 \texttimes{} 512 &  60 M &  939 G & 23.3 & 46.7 \\
        \nb\textbf{NAT-T}                       & 2048 \texttimes{} 512 &  58 M &  934 G & 21.4 & 48.4 \\
        \ours \textbf{DiNAT-T}                  & 2048 \texttimes{} 512 &  58 M &  934 G & 21.3 & \textbf{48.8} \\
        \midrule
        \wb\textbf{Swin-S}                      & 2048 \texttimes{} 512 &  81 M & 1040 G & 17.0 & 49.5 \\
        \cb\textbf{ConvNeXt-S}                  & 2048 \texttimes{} 512 &  82 M & 1027 G & 19.1 & 49.6 \\
        \nb\textbf{NAT-S}                       & 2048 \texttimes{} 512 &  82 M & 1010 G & 17.9 & 49.5 \\
        \ours \textbf{DiNAT-S}                  & 2048 \texttimes{} 512 &  82 M & 1010 G & 18.1 & \textbf{49.9} \\
        \midrule
        \wb\textbf{Swin-B}                      & 2048 \texttimes{} 512 & 121 M & 1188 G & 14.6 & 49.7 \\
        \cb\textbf{ConvNeXt-B}                  & 2048 \texttimes{} 512 & 122 M & 1170 G & 16.4 & 49.9 \\
        \nb\textbf{NAT-B}                       & 2048 \texttimes{} 512 & 123 M & 1137 G & 15.6 & 49.7 \\
        \ours \textbf{DiNAT-B}                  & 2048 \texttimes{} 512 & 123 M & 1137 G & 15.4 & \textbf{50.4} \\
        \midrule
        \wb\textbf{Swin-L\dgr\dgg}              & 2560 \texttimes{} 640 & 234 M & 2585 G &  8.5 & 53.5 \\
        \cb\textbf{ConvNeXt-L\dgg}              & 2560 \texttimes{} 640 & 235 M & 2458 G &  9.6 & 53.7 \\
        \ours \textbf{DiNAT-L\dgg}              & 2560 \texttimes{} 640 & 238 M & 2335 G &  9.0 & \textbf{54.9} \\
        \bottomrule
    \end{tabular}
    }
    \caption{
    \textbf{ADE20K semantic segmentation performance.} 
    \dgg indicates that the model was pre-trained on ImageNet-22K. 
    \dgr indicates increased window size from the default 7\sq{} to 12\sq{}. 
    \downstreamnotes{}
    }
    \label{tab:semseg}
\end{table}

We also trained UPerNet~\cite{xiao2018unified} with our DiNAT as the backbone on ADE20K~\cite{zhou2017scene}, with ImageNet-pre-trained backbones. 
We followed NAT's \verb|mmsegmentation|~\cite{mmseg2020} (Apache License v2) configurations, itself following Swin's configuration for training ADE20K.
The results are presented in \cref{tab:semseg}.
We find that DiNAT exhibits a noticeable improvement over the original NAT model.
DiNAT also maintains its place ahead of both models at scale with ImageNet-22K pre-training.

\subsection{Ablation study}
\label{subsec:miscexps}
In this section, we aim to study DiNAT in more depth by analyzing the effects of: dilation values, NA-DiNA order, kernel sizes, and test-time changes in dilation.

\paragraph{Dilation values. } In ~\cref{tab:dilationperformance}, we present models with different dilation values, and their effect on classification, detection, instance segmentation and semantic segmentation performance levels.
Note that the increased dilation (16, 8, 4, 2) is applicable to downstream tasks only, because in theory input feature maps should be larger than or equal to the product of kernel size and dilation. As a result, ``8, 4, 2, 1'' is the maximum applicable dilation to ImageNet at 224 \texttimes{} 224 resolution. Depending on image resolution, even higher dilation values are possible. We explored a ``dynamic'' dilation value, where DiNA layers apply the maximum possible dilation, which is the floor of resolution divided by kernel size (``Maximum'' in \cref{tab:dilationperformance}). We finally choose settle on ``gradual'' dilation (see illustration in \cref{fig:connections}), in which we gradually increase dilation to the maximum level defined. For instance, if maximum dilation for a specific level to 8, its layers will have dilation values 1, 2, 1, 4, 1, 6, 1, 8 (refer to \cref{appsec:hyperparameters} for details).

\setlength{\tabcolsep}{6pt}
\begin{table}[t]
    \centering
    \resizebox{0.475\textwidth}{!}{
    \begin{tabular}{lc|c|cc|c}
        \toprule
        \textbf{Model}                              & \textbf{Dilation}             & \textbf{ImageNet} & \multicolumn{2}{c}{\textbf{MSCOCO}}                               & \textbf{ADE20K}   \\
                                                    & \textbf{per level}            & \textbf{Top-1 (\%)}& \textbf{AP\textsuperscript{b}} & \textbf{AP\textsuperscript{m}}   & \textbf{mIoU}     \\
        \midrule
        \nb\textbf{NAT-Tiny}                                & 1, 1, 1, 1                    & 83.2    & 47.7                           & 42.6                             & 48.4              \\
        \db\textbf{DiNAT-Tiny}                              & 8, 4, 2, 1                    & 82.7    & 48.0                           & 42.9                             & 48.5              \\
        \db\textbf{DiNAT-Tiny}                              & 16, 8, 4, 2                   & -       & 48.3                           & 43.4                             & 48.5              \\
        \db\textbf{DiNAT-Tiny}                              & Maximum                       & 82.7    & \textbf{48.6}                  & \textbf{43.5}                    & 48.7              \\
        \ours \textbf{DiNAT-Tiny}                           & Gradual                       & -       & \textbf{48.6}                  & \textbf{43.5}                    & \textbf{48.8}     \\
        \bottomrule
    \end{tabular}
    }
    \caption{
    \textbf{Dilation impact on performance.} 
    Dilation values beyond "8, 4, 2, 1" are only applicable to downstream tasks, as their larger resolution allows for it.
    Maximum dilation indicates it is set to the maximum possible value based on input size. It would be the same as "8, 4, 2, 1" for ImageNet.
    Gradual dilation indicates that dilation values in DiNA layers increase gradually.
    }
    \label{tab:dilationperformance}
\end{table}

\paragraph{NA-DiNA \vs  DiNA-NA. } We also experimented with models with DiNA layers before NA layers, as opposed to our final NA before DiNA choice.
While the local-global order (NA-DiNA) was our initial choice, we've also found it to be the more effective choice.
We also tried a model with only DiNA modules, and found that it performs significantly worse than other combinations. 
This highlights the importance of having a combination of both local and sparse global attention patterns in the model.
The results are summarized in \cref{tab:nadinaorderperformance}. 

\setlength{\tabcolsep}{4pt}
\begin{table}[t]
    \centering
    \resizebox{0.475\textwidth}{!}{
    \begin{tabular}{lc|c|cc|c}
        \toprule
        \textbf{Variant}                                & \textbf{Layer}            & \textbf{ImageNet} & \multicolumn{2}{c|}{\textbf{MSCOCO}}                              & \textbf{ADE20K}   \\
                                                        & \textbf{structure}        & \textbf{Top-1 (\%)}& \textbf{AP\textsuperscript{b}} & \textbf{AP\textsuperscript{m}}   & \textbf{mIoU}     \\
        \midrule
        \nb\nb\textbf{NAT-Tiny}                         & NA-NA                     & \textbf{83.2}   & 47.7                           & 42.6                             & 48.4              \\
        \grayrow\nb\db\textbf{DiNAT-Tiny}               & NA-DiNA                   & 82.7            & 48.3                           & 43.4                             & \textbf{48.5}     \\
        \db\nb                                          & DiNA-NA                   & 82.6            & \textbf{48.5}                  & \textbf{43.5}                    & 47.9              \\
        \db\db                                          & DiNA-DiNA                 & 82.2            & 44.9                           & 40.5                             & 45.8              \\
        \bottomrule
    \end{tabular}
    }
    \caption{
    \textbf{Layer structure impact on performance.} 
    Our final model has the local-global (NA-DiNA) order.
    }
    \label{tab:nadinaorderperformance}
\end{table}

\paragraph{Kernel size. } We study the effect of kernel size on model performance in \cref{tab:kernelsize}. 
We observed that a DiNAT-Tiny sees a significant decay in performance with a smaller kernel size across all three tasks.
However, we find increasing kernel size beyond the default 7\texttimes{}7 does not result in a significant increase in return.
\setlength{\tabcolsep}{3pt}
\begin{table}[t]
    \centering
    \resizebox{0.475\textwidth}{!}{
    \begin{tabular}{ll|cc|ccc|cc}
        \toprule
        \textbf{Model}  & \textbf{Win.}         & \multicolumn{2}{c|}{\textbf{ImageNet}} & \multicolumn{3}{c|}{\textbf{MSCOCO}}                                               & \multicolumn{2}{c}{\textbf{ADE20K}}  \\
                        & \textbf{size}           & \textbf{Top-1}& \textbf{Thru.}     & \textbf{AP\textsuperscript{b}} & \textbf{AP\textsuperscript{m}}   & \textbf{Thru.} & \textbf{mIoU} & \textbf{Thru.}       \\
        \midrule
        \nb\textbf{NAT-T}               & 5\sq{}          & \textbf{81.6}   & 1810 \ips         & 46.8                           & 42.0                             & 45.5 \fps      & 46.3          & 22.9 \fps            \\
        \ours \textbf{DiNAT-T}          & 5\sq{}          & 81.3            & 1777 \ips         & \textbf{47.6}                  & \textbf{42.7}                    & 45.6 \fps      & \textbf{46.4} & 22.7 \fps            \\
        \midrule
        \nb\textbf{NAT-T}               & 7\sq{}          & \textbf{83.2}   & 1537 \ips         & 47.7                           & 42.6                             & 44.5 \fps      & 48.4          & 21.4 \fps            \\
        \ours \textbf{DiNAT-T}          & 7\sq{}          & 82.7            & 1500 \ips         & \textbf{48.3}                  & \textbf{43.4}                    & 43.3 \fps      & \textbf{48.5} & 21.3 \fps            \\
        \midrule
        \nb\textbf{NAT-T}               & 9\sq{}          & \textbf{83.1}   & 1253 \ips         & 48.5                           & 43.3                             & 39.4 \fps      & 48.1          & 20.2 \fps            \\
        \ours \textbf{DiNAT-T}          & 9\sq{}          & \textbf{83.1}   & 1235 \ips         & \textbf{48.8}                  & \textbf{43.5}                    & 39.2 \fps      & \textbf{48.4} & 20.0 \fps            \\
        \bottomrule
    \end{tabular}
    }
    \caption{
    \textbf{Kernel size impact on performance.} 
    Note that we set dilation to the maximum values possible in each block based on the default resolutions. 
    Therefore, the variant with kernel size 5 has larger dilation values compared to the one with kernel size 7. 
    }
    \label{tab:kernelsize}
\end{table}

\paragraph{Test-time dilation changes. } We present an analysis of sensitivity to dilation values, in which we attempt different dilation values on already trained models, and evaluate their performance.
This can be particularly important to cases with varying resolutions, i.e. multi-scale testing. 
For DiNAT to be at its best, dilation level needs to be a near-maximum number to expand attention to a longer range.
The results are presented in \cref{tab:testtimedilationchange}.
\setlength{\tabcolsep}{3pt}
\begin{table}[t]
    \centering
    \resizebox{0.475\textwidth}{!}{
    \begin{tabular}{lcc|c|cc|c}
        \toprule
        \textbf{Model}      & \multicolumn{2}{c|}{\textbf{Dilation}}                   & \textbf{ImageNet} & \multicolumn{2}{c|}{\textbf{MSCOCO}}                               & \textbf{ADE20K}   \\
                            & \textbf{Train}              & \textbf{Test}             & \textbf{Top-1 (\%)}& \textbf{AP\textsuperscript{b}} & \textbf{AP\textsuperscript{m}}   & \textbf{mIoU}     \\
        \midrule
        \oursn\textbf{NAT-T}            & 1, 1, 1, 1                  & 1, 1, 1, 1                & 83.2            & 47.7                           & 42.6                             & 48.4              \\
                                        & 1, 1, 1, 1                  & 8, 4, 2, 1                & 81.0            & 42.6                           & 39.5                             & 46.3              \\
                                        & 1, 1, 1, 1                  & 16, 8, 4, 2               & -               & 36.0                           & 34.4                             & 40.2              \\
                                        & 1, 1, 1, 1                  & Maximum                   & -               & 31.7                           & 30.7                             & 38.2              \\
        \midrule
                                        & 8, 4, 2, 1                  & 1, 1, 1, 1                & 78.2            & 43.0                           & 38.6                             & 41.5              \\
        \ours\textbf{DiNAT-T}           & 8, 4, 2, 1                  & 8, 4, 2, 1                & 82.7            & 48.0                           & 42.9                             & 48.5              \\
                                        & 8, 4, 2, 1                  & 16, 8, 4, 2               & -               & 45.6                           & 41.3                             & 47.1              \\
                                        & 8, 4, 2, 1                  & Maximum                   & -               & 40.2                           & 37.3                             & 45.8              \\
        \midrule
                                        & 16, 8, 4, 2                 & 1, 1, 1, 1                & -               & 29.0                           & 26.7                             & 26.2              \\
                                        & 16, 8, 4, 2                 & 8, 4, 2, 1                & -               & 42.6                           & 38.6                             & 43.3              \\
        \ours\textbf{DiNAT-T}           & 16, 8, 4, 2                 & 16, 8, 4, 2               & -               & 48.3                           & 43.4                             & 48.5              \\
                                        & 16, 8, 4, 2                 & Maximum                   & -               & 47.4                           & 42.5                             & 48.6              \\
        \bottomrule
    \end{tabular}
    }
    \caption{
    \textbf{Test time dilation change and its impact on performance.} 
    Dilation values larger than 8, 4, 2, 1 are inapplicable to ImageNet at 224\sq{}. 
    }
    \label{tab:testtimedilationchange}
\end{table}

\subsection{Image segmentation with Mask2Former}
\label{sec:m2f}

To analyze DiNAT's segmentation performance further, we conducted experiments with Mask2Former~\cite{cheng2022masked}.
Mask2Former is an attention-based segmentation architecture, which can be trained on instance segmentation, semantic segmentation, and panoptic segmentation.
It set a new state-of-the-art score for panoptic and instance segmentation on MS-COCO, as well as semantic segmentation on ADE20K.
Mask2Former additionally used Swin-Large as the backbone, making it the perfect candidate for this experiment.
We trained Mask2Former on MS-COCO~\cite{lin2014microsoft}, ADE20K~\cite{zhou2017scene}, and Cityscapes~\cite{cordts2016cityscapes}, on all three segmentation objectives (instance, semantic, and panoptic), by simply replacing the Swin-Large backbone in a fork of their original repository.
Following their reported environment, we used PyTorch 1.9 with Detectron2~\cite{wu2019detectron2}.
We present instance segmentation results in \cref{apptab:mask2formerinsseg}, semantic segmentation results in \cref{apptab:mask2formersemseg}, and panoptic segmentation results in \cref{apptab:mask2formerpanseg}.
We note that DiNAT-L is using an 11\sq{} kernel size, instead of Swin-L's 12\sq{}, since even-sized windows break the symmetry in NA and are therefore not defined.

DiNAT-L outperforms Swin-L on all three tasks and datasets. 
It also sets new state of the art records for image segmentation without using extra data.
According to PapersWithCode leaderboards, DiNAT-L with Mask2Former is the SOTA panoptic segmentation on ADE20K and MS-COCO, and instance segmentation on ADE20K and Cityscapes.
It also ties with the current SOTA on ADE20K, and ranks second on Cityscapes semantic segmentation (previous SOTA on both is SeMask~\cite{jain2021semask}).

\setlength{\tabcolsep}{2pt}
\begin{table}[t]
    \centering
    \resizebox{0.475\textwidth}{!}{
    \begin{tabular}{l|ccc|ccccc}
        \toprule
        \textbf{Backbone} & \textbf{Win.} & \textbf{\# of}     & \textbf{FLOPs} & \textbf{AP} & \textbf{AP\textsuperscript{50}} & \textbf{AP\textsuperscript{S}} & \textbf{AP\textsuperscript{M}} & \textbf{AP\textsuperscript{L}}\\ 
                          & \textbf{Size} & \textbf{Params}    &                & &  \\ 
        \midrule
        \multicolumn{9}{c}{\textit{MS-COCO}} \\
        \midrule
        \wb\textbf{Swin-L}              & 12 \texttimes{} 12 & 216 M & 641 G & 50.1 & - & 29.9 & 53.9 & \textbf{72.1} \\
        \ours \textbf{DiNAT-L}          & 11 \texttimes{} 11 & 220 M & 522 G & \textbf{50.8} & \textbf{75.0} & \textbf{30.9} & \textbf{54.7} & \textbf{72.1} \\
        \midrule
        \multicolumn{9}{c}{\textit{ADE20K}} \\
        \midrule
        \wb\textbf{Swin-L}              & 12 \texttimes{} 12 & 216 M & 654 G & 34.9 & - & \textbf{16.3} & \textbf{40.0} & 54.7\\
        \ours \textbf{DiNAT-L}          & 11 \texttimes{} 11 & 220 M & 535 G & \textbf{35.4} & - & \textbf{16.3} & 39.0 & \textbf{55.5}\\
        \midrule
        \multicolumn{9}{c}{\textit{Cityscapes}} \\
        \midrule
        \wb\textbf{Swin-L}              & 12 \texttimes{} 12 & 216 M & 641 G & 43.7 & 71.4 & - & - \\
        \ours \textbf{DiNAT-L}          & 11 \texttimes{} 11 & 220 M & 522 G & \textbf{45.1} & \textbf{72.6} & - & - & - \\
        \bottomrule
    \end{tabular}
    }
    \caption{
    \textbf{Instance segmentation performance with Mask2Former.} 
    All backbones were pre-trained on ImageNet-22K. 
    FLOPs are reported with respect to resolution 800\sq{}.
    }
    \label{apptab:mask2formerinsseg}
\end{table}

\setlength{\tabcolsep}{5pt}
\begin{table}[t]
    \centering
    \resizebox{0.475\textwidth}{!}{
    \begin{tabular}{l|ccc|cc}
        \toprule
        \textbf{Backbone} & \textbf{Win.} & \textbf{\# of}     & \textbf{FLOPs} & \multicolumn{2}{c}{\textbf{mIoU}}\\ 
                          & \textbf{Size} & \textbf{Params}    &                & single scale & multi scale \\ 
        \midrule
        \multicolumn{6}{c}{\textit{ADE20K}} \\
        \midrule
        \wb\textbf{Swin-L}              & 12 \texttimes{} 12 & 215 M & 636 G & 56.1 & 57.3 \\
        \ours \textbf{DiNAT-L}          & 11 \texttimes{} 11 & 220 M & 518 G & \textbf{57.3} & \textbf{58.1}  \\
        \midrule
        \multicolumn{6}{c}{\textit{Cityscapes}} \\
        \midrule
        \wb\textbf{Swin-L}              & 12 \texttimes{} 12 & 215 M & 627 G & 83.3 & 84.3 \\
        \ours \textbf{DiNAT-L}          & 11 \texttimes{} 11 & 220 M & 509 G & \textbf{83.9} & \textbf{84.5}  \\
        \bottomrule
    \end{tabular}
    }
    \caption{
    \textbf{Semantic segmentation performance with Mask2Former.} 
    All backbones were pre-trained on ImageNet-22K. 
    FLOPs are reported with respect to resolution 800\sq{}.
    }
    \label{apptab:mask2formersemseg}
\end{table}

\setlength{\tabcolsep}{2pt}
\begin{table}[t]
    \centering
    \resizebox{0.475\textwidth}{!}{
    \begin{tabular}{l|ccc|ccc|cc}
        \toprule
        \textbf{Backbone} & \textbf{Win.} & \textbf{\# of}     & \textbf{FLOPs} & \textbf{PQ} & \textbf{PQ\textsuperscript{Th}} & \textbf{PQ\textsuperscript{St}} & \textbf{AP$^{\text{Th}}_{\text{pan}}$} & \textbf{mIoU\textsubscript{pan}} \\ 
                          & \textbf{Size} & \textbf{Params}    &                & &  \\ 
        \midrule
        \multicolumn{9}{c}{\textit{MS-COCO}} \\
        \midrule
        \wb\textbf{Swin-L}              & 12 \texttimes{} 12 & 216 M & 658 G & 57.8 & 64.2 & 48.1 & 48.6 & 67.4 \\
        \ours \textbf{DiNAT-L}          & 11 \texttimes{} 11 & 220 M & 540 G & \textbf{58.5} & \textbf{64.9} & \textbf{48.8} & \textbf{49.2} & \textbf{68.3} \\
        \midrule
        \multicolumn{9}{c}{\textit{ADE20K}} \\
        \midrule
        \wb\textbf{Swin-L}              & 12 \texttimes{} 12 & 216 M & 660 G & 48.1 & - & - & 34.2 & 54.5 \\
        \ours \textbf{DiNAT-L}          & 11 \texttimes{} 11 & 220 M & 542 G & \textbf{49.4} & - & - & \textbf{35.0} & \textbf{56.3} \\
        \midrule
        \multicolumn{9}{c}{\textit{Cityscapes}} \\
        \midrule
        \wb\textbf{Swin-L}              & 12 \texttimes{} 12 & 216 M & 643 G & 66.6 & - & - & 43.6 & 82.9 \\
        \ours \textbf{DiNAT-L}          & 11 \texttimes{} 11 & 220 M & 525 G & \textbf{67.2} & - & - & \textbf{44.5} & \textbf{83.4} \\
        \bottomrule
    \end{tabular}
    }
    \caption{
    \textbf{Panoptic segmentation performance with Mask2Former.} 
    All backbones were pre-trained on ImageNet-22K. 
    FLOPs are reported with respect to resolution 800\sq{}.
    }
    \label{apptab:mask2formerpanseg}
\end{table}

\section{Conclusion}
\label{sec:conclusion}
Local attention modules are effective at reducing complexity, and are crucial when working with a hierarchical model that gradually downsamples inputs.
Nevertheless, they cannot capture longer range inter-dependencies as well as global self attention, unless their receptive field size is increased, which defeats their initial purpose of efficiency and tractability.
In this paper, we propose DiNA, a natural extension to NA that expands its local attention to sparse global attention at no additional cost. 
We build DiNAT with combinations of NA and DiNA, and show that it can improve performance significantly, especially in downstream tasks, without introducing any additional computational burden.
Paired with new segmentation frameworks, our model achieves state-of-the-art image semantic, instance, and panoptic segmentation performance
While our experiments give insight into the power behind such flexible attention modules, neither their performance nor efficiency stop here. 
We believe that combinations of NA and DiNA will be able to empower various models in vision and beyond, wherever locality and global context matter.
We open source our entire project, including our extension to \natten{}, and will continue to support it as a toolkit for the community to allow easy experimentation with sparse sliding-window attention.

\paragraph{Acknowledgments.}
We thank Picsart AI Research (PAIR), Meta/Facebook AI, and Intelligence Advanced Research Projects Activity (IARPA) for their generous support that made this work possible.

{\small
\bibliographystyle{ieee_fullname}
\bibliography{references}
}

\clearpage \appendix
\renewcommand{\thetable}{\Roman{table}}
\renewcommand{\thefigure}{\Roman{figure}}
\setcounter{table}{0}
\setcounter{figure}{0}

\begin{figure*}[t]
    \centering
    \begin{subfigure}{0.4\textwidth}
        \centering
        \includegraphics[trim={6mm, 6mm, 5mm, 6mm},clip,width=\textwidth]{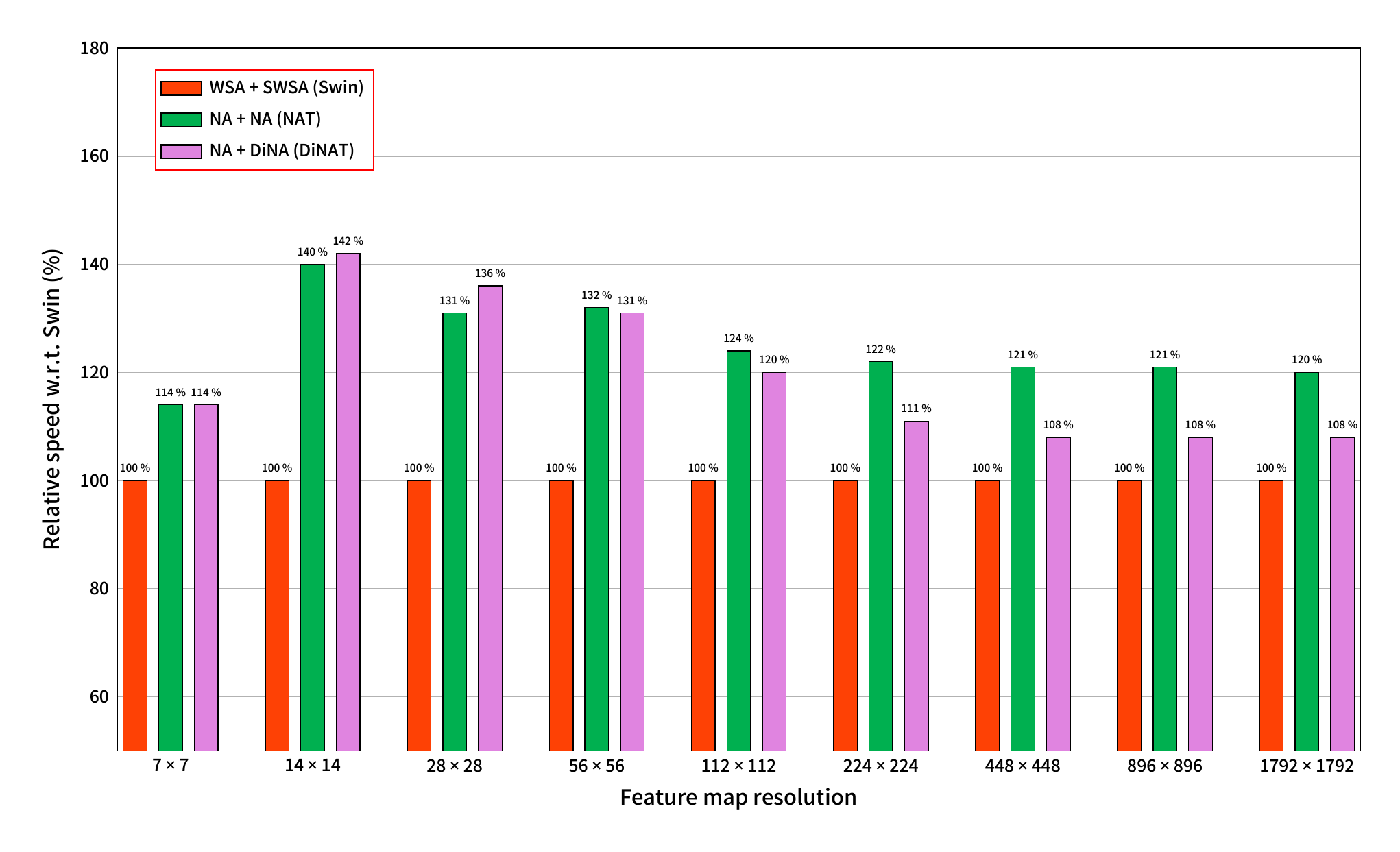}
        \caption{Speed.}
        \label{appfig:cudatime}
    \end{subfigure}
    \hfill
    \begin{subfigure}{0.595\textwidth}
        \centering
        \includegraphics[trim={6mm, 6mm, 5mm, 5mm},clip,width=\textwidth]{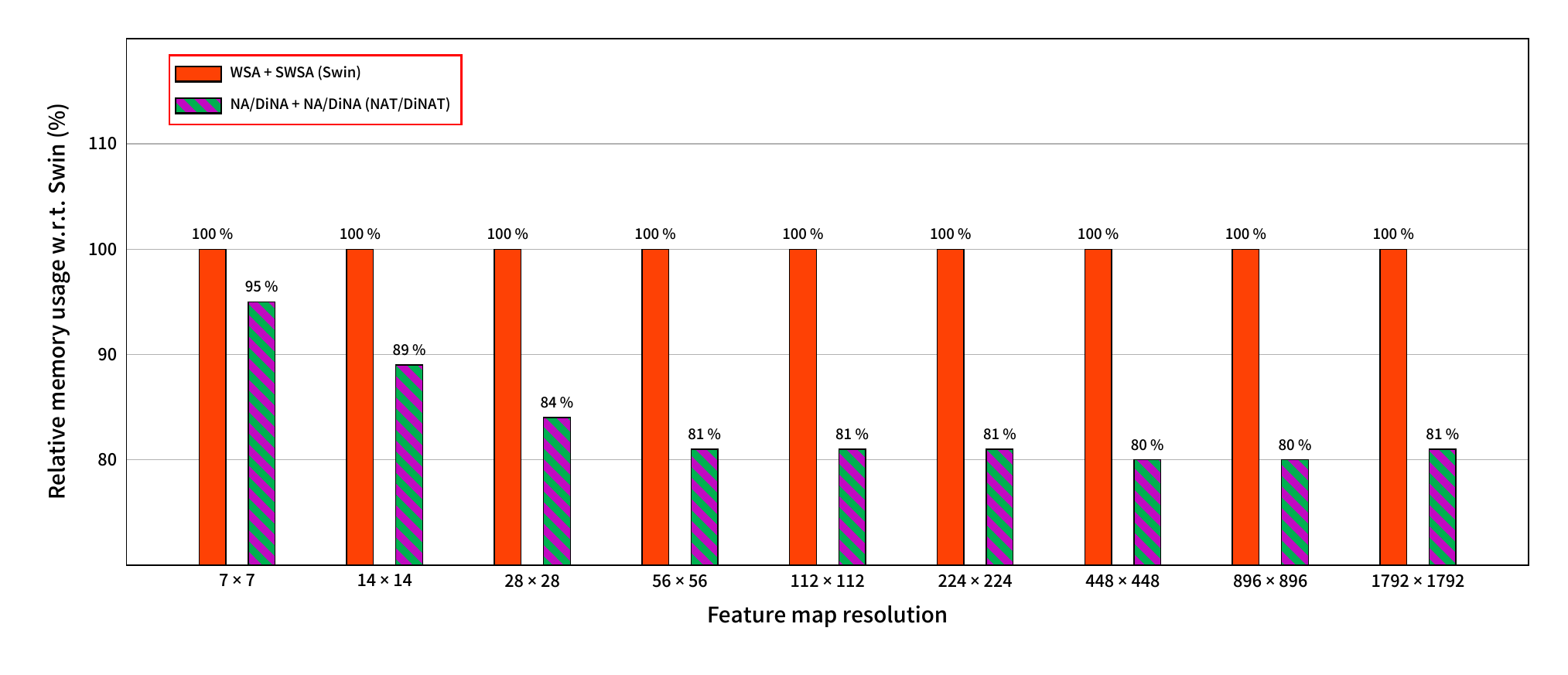}
        \caption{Memory usage.}
        \label{appfig:cudamemory}
    \end{subfigure}
    \caption{
    \textbf{Layer-wise relative speed and memory comparison between NAT and DiNAT, with respect to Swin.} 
    NAT layers, which are only two consecutive NA layers with kernel size 7\sq{}, are already up to 40\% faster than Swin layers with the same kernel size.
    DiNAT layers, comprised of an NA layer followed by a DiNA layer, are slightly slower in practice due to the break in memory access pattern, but are still faster than Swin layers.
    }
    \label{appfig:cudatimeandmemory}
\end{figure*}

\section*{\Large{Appendix}}

\section{Implementation notes}
\label{appsec:implementation}

As discussed in \cref{subsec:implementation}, we extend the existing \natten{} package to support dilated neighborhoods.
\natten{} has a two-stage attention computation, similar to many other implementations: QK, and AV. The former computes the dot product of queries and keys, and produces attention weights, and the latter applies attention weights to the values. Scaling, softmax, and dropout are not included, as to prevent re-implementation.
One of the advantages of this two-stage structure over manual implementations is that, like implementations of convolutions, sliding windows are taken directly from the source tensor, and not cached into an intermediary tensor, thus using significantly less memory. We refer readers to \natten{} documentation, and NAT~\cite{hassani2022neighborhood} for further details.

\paragraph{Dilation support.} Adding dilation to \natten{}'s naive kernels is mostly simple: instead of incrementing neighbors across each axis by $1$, we simply instruct the kernels to increment by a variable $d$. NA however has a special way to handle edge/corner pixels, which requires additional changes to support dilation.
The greater challenge in adding dilation to \natten{} was adding it to the ``tiled'' kernels that utilize shared memory.
Tiled NA kernels are a more recent addition to \natten{}, and boost NA's throughput significantly. 
Tiled implementations of matrix multiplication and convolutions are essential in parallelizing these operations efficiently, while minimizing DRAM accesses.
As the name suggests, tiled implementations divide the operation into tiles and cache tiles o inputs from the global memory into the shared memory within each threadblock.
Accessing values from shared memory is typically much faster compared to directly accessing global memory, but also comes with challenges such as bank conflicts.
Tiled implementations also operate with the assumption that access patterns are not broken.
Introducing dilation values would break those access patterns and require a re-implementation that ensures dilated neighbors are cached instead of local neighbors.
We present a layer-wise relative speed and memory usage comparison between NAT and DiNAT with respect to Swin in \cref{appfig:cudatimeandmemory}.

\paragraph{Scaling and brain float support.} In order to train our larger models and avoid overflowing activation values in later layers of the model, we've had to switch from automatic mixed-precision training with the default half precision data type, float16, which has 5 exponent bits and 10 mantissa bits, to bfloat16, which has the advantage of having 8 exponent bits while having only 7 mantissa bits. Utilizing bfloat16 has often been recommended for cases which lead to large activations, which includes ours as we scale our model. However, switching to bfloat16 required a re-implementation of \natten{}'s half precision kernels to support and utilize bfloat16 correctly.

\section{Training settings}
\label{appsec:hyperparameters}
We provide additional details on training DiNAT in \cref{apptab:variants}.
We also provide details on DiNAT$_s$, which utilizes non-overlapping patch embedding and downsampling, similar to Swin~\cite{liu2021swin} and ConvNeXt~\cite{liu2022convnet}.
DiNAT$_s$ serves as an alternative DiNA-based model, which has an architecture identical to Swin.
DiNAT$_s$ can also serve as an ablation model, since it is identical to Swin in architecture, with WSA replaced with NA, and SWSA replaced with DiNA.

\setlength{\tabcolsep}{3pt}
\begin{table}[t]
    \centering
    \resizebox{0.475\textwidth}{!}{
    \begin{tabular}{lcccccc}
        \toprule
        \textbf{Variant}                        & \textbf{Downsampling} & \textbf{Layers}   & \textbf{Dim \texttimes{}}         & \textbf{MLP}          & \textbf{\# of}        & \textbf{FLOPs}\\
                                                &                       & \textbf{per level}& \textbf{Heads}                    & \textbf{ratio}        & \textbf{Params}       &               \\
        \midrule
        \db\textbf{DiNAT$_s$-T}         & Patched               & 2, 2,  6, 2       & 32 \texttimes{} 3                 & 4                     &  28 M                 &  4.5 G \\
        \db\textbf{DiNAT$_s$-S}         & Patched               & 2, 2, 18, 2       & 32 \texttimes{} 3                 & 4                     &  50 M                 &  8.7 G \\
        \db\textbf{DiNAT$_s$-B}         & Patched               & 2, 2, 18, 2       & 32 \texttimes{} 4                 & 4                     &  88 M                 & 15.4 G \\
        \db\textbf{DiNAT$_s$-L}         & Patched               & 2, 2, 18, 2       & 32 \texttimes{} 6                 & 4                     & 197 M                 & 34.5 G \\
        \midrule
        \db\textbf{DiNAT-M}             & Conv                  & 3, 4,  6, 5       & 32 \texttimes{} 2                 & 3                     &  20 M                 &  2.7 G \\
        \db\textbf{DiNAT-T}             & Conv                  & 3, 4, 18, 5       & 32 \texttimes{} 2                 & 3                     &  28 M                 &  4.3 G \\
        \db\textbf{DiNAT-S}             & Conv                  & 3, 4, 18, 5       & 32 \texttimes{} 3                 & 2                     &  51 M                 &  7.8 G \\
        \db\textbf{DiNAT-B}             & Conv                  & 3, 4, 18, 5       & 32 \texttimes{} 4                 & 2                     &  90 M                 & 13.7 G \\
        \db\textbf{DiNAT-L}             & Conv                  & 3, 4, 18, 5       & 32 \texttimes{} 6                 & 2                     & 200 M                 & 30.6 G \\
        \bottomrule
    \end{tabular}
    }
    \caption{\textbf{Summary of DiNAT and DiNAT$_s$ configurations.} 
    Channels (heads and dim) double after every level until the final one.
    Default dilation values for the four levels are 8, 4, 2, and 1. Kernel size is 7\texttimes{}7 in all variants.
    }
    \label{apptab:variants}
\end{table}

\setlength{\tabcolsep}{4pt}
\begin{table}[t]
    \centering
    \resizebox{0.475\textwidth}{!}{
    \begin{tabular}{lc|cccc}
        \toprule
        \textbf{Variant}                  & \textbf{Resolution} & \textbf{Level 1} & \textbf{Level 2}       & \textbf{Level 3}          & \textbf{Level 4}         \\
        \midrule
        \multicolumn{6}{c}{\textit{ImageNet classification.}}\\
        \midrule
        \db\textbf{DiNAT$_s$-T}     & 224\sq  & 1, 8             & 1, 4                   & 1, 2, 1, 2, 1, 2 & 1, 1   \\
        \db\textbf{DiNAT$_s$-S/B/L} & 224\sq  & 1, 8             & 1, 4                   & 1, 2, 1, 2, 1, 2, 1, 2, 1, 2, 1, 2, 1, 2, 1, 2, 1, 2  & 1, 1   \\
        \db\textbf{DiNAT$_s$-L}     & 384\sq  & 1, 13            & 1, 6                   & 1, 3, 1, 3, 1, 3, 1, 3, 1, 3, 1, 3, 1, 3, 1, 3, 1, 3  & 1, 1   \\
        \midrule
        \db\textbf{DiNAT-M}         & 224\sq  & 1, 8, 1          & 1, 4, 1, 4 & 1, 2, 1, 2, 1, 2 & 1, 1, 1, 1, 1   \\
        \db\textbf{DiNAT-T/S/B/L}   & 224\sq  & 1, 8, 1          & 1, 4, 1, 4 & 1, 2, 1, 2, 1, 2, 1, 2, 1, 2, 1, 2, 1, 2, 1, 2, 1, 2  & 1, 1, 1, 1, 1   \\
        \db\textbf{DiNAT-L}         & 384\sq  & 1, 13, 1         & 1, 6, 1, 6 & 1, 3, 1, 3, 1, 3, 1, 3, 1, 3, 1, 3, 1, 3, 1, 3, 1, 3  & 1, 1, 1, 1, 1   \\
        \midrule
        \multicolumn{6}{c}{\textit{MS-COCO detection and instance segmentation.}}\\
        \midrule
        \db\textbf{DiNAT$_s$-T}     & 800\sq  & 1, 28             & 1, 14                  & 1, 3, 1, 5, 1, 7  & 1, 3   \\
        \db\textbf{DiNAT$_s$-S/B/L} & 800\sq  & 1, 28             & 1, 14                  & 1, 3, 1, 5, 1, 7, 1, 3, 1, 5, 1, 7, 1, 3, 1, 5, 1, 7  & 1, 3   \\
        \midrule
        \db\textbf{DiNAT-M}         & 800\sq  & 1, 28, 1          & 1, 7, 1, 14            & 1, 3, 1, 5, 1, 7  & 1, 3, 1, 3, 1 \\
        \db\textbf{DiNAT-T/S/B/L}   & 800\sq  & 1, 28, 1          & 1, 7, 1, 14            & 1, 3, 1, 5, 1, 7, 1, 3, 1, 5, 1, 7, 1, 3, 1, 5, 1, 7  & 1, 3, 1, 3, 1 \\
        \midrule
        \multicolumn{6}{c}{\textit{ADE20K semantic segmentation.}}\\
        \midrule
        \db\textbf{DiNAT$_s$-T}     & 512\sq  & 1, 16             & 1, 8                   & 1, 2, 1, 3, 1, 4  & 1, 2   \\
        \db\textbf{DiNAT$_s$-S/B}   & 512\sq  & 1, 16             & 1, 8                   & 1, 2, 1, 3, 1, 4, 1, 2, 1, 3, 1, 4, 1, 2, 1, 3, 1, 4 & 1, 2   \\
        \db\textbf{DiNAT$_s$-L}     & 640\sq  & 1, 20             & 1, 10                  & 1, 2, 1, 3, 1, 4, 1, 5, 1, 2, 1, 3, 1, 4, 1, 5, 1, 5 & 1, 2   \\
        \midrule
        \db\textbf{DiNAT-M}         & 512\sq  & 1, 16, 1          & 1, 4, 1, 8             & 1, 2, 1, 3, 1, 4  & 1, 2, 1, 2, 1 \\
        \db\textbf{DiNAT-T/S/B}     & 512\sq  & 1, 16, 1          & 1, 4, 1, 8             & 1, 2, 1, 3, 1, 4, 1, 2, 1, 3, 1, 4, 1, 2, 1, 3, 1, 4  & 1, 2, 1, 2, 1 \\
        \db\textbf{DiNAT-L}         & 640\sq  & 1, 20, 1          & 1, 5, 1, 10            & 1, 2, 1, 3, 1, 4, 1, 5, 1, 2, 1, 3, 1, 4, 1, 5, 1, 5  & 1, 2, 1, 2, 1 \\
        \bottomrule
    \end{tabular}
    }
    \caption{\textbf{Dilation values.} 
    Due to ImageNet's relatively small input resolution, level 4 layers cannot go beyond a dilation value of 1, which is equivalent to NA. 
    Also note that at 224\texttimes{}224 resolution, level 4 inputs will be exactly 7\texttimes{}7, therefore NA will be equivalent to self attention. 
    This is not true in downstream tasks where resolutions are noticeably higher where levels 2 and 3 have \textit{gradually} increasing dilation values, which are repeated in deeper models. 
    This corresponds to the highlighted rows in \cref{tab:dilationperformance} labeled ``Gradual''. 
    These configurations apply to all downstream experiments (excluding those in \cref{subsec:miscexps}).
    }
    \label{apptab:dinatsettings}
\end{table}

One of the most important architecture-related hyperparameters in DiNA-based models is dilation values. 
Both DiNAT and DiNAT$_s$ use a combination of NA and DiNA layers.
We typically set dilation values in DiNA layers to be the maximum possible value with respect to input resolutions, if known.
For example, ImageNet classification at 224\texttimes{}224 is downsampled to a quarter of the original size initially, therefore Level 1 layers take feature maps of resolution 56\texttimes{}56 as input.
With a kernel size of 7\texttimes{}7, the maximum possible dilation value is $\lfloor 56 / 7 \rfloor = 8$. Level 2 will take feature maps of resolution 28\texttimes{}28 as input, leading to a maximum possible dilation value of $4$.
Because of this, we change dilation values depending on the task and resolution.
We present the final dilation values we used in classification, detection, and segmentation in \cref{apptab:dinatsettings}.
Note that we only change dilation values for DiNA layers, since we found that fine-tuning NA layers to DiNA layers may result in a slight decrease in initial performance (see \cref{subsec:miscexps}, \cref{tab:testtimedilationchange}).

\section{Experiments with alternative architecture}

\label{appsec:altarchresults}
We conducted all primary experiments with both our main model, DiNAT, as well as DiNAT$_s$.
We found that DiNAT$_s$ could serve as alternatives in certain cases, as they still provide noticeable improvements over Swin in terms of speed, accuracy, and memory usage.
Classification results are provided in \cref{apptab:imagenet_comparison}, object detection and instance segmentation results are provided in \cref{apptab:objectdetection}, and semantic segmentation results are provided in \cref{apptab:semseg}.

\setlength{\tabcolsep}{3pt}
\begin{table}[t]
    \centering
    \resizebox{0.475\textwidth}{!}{
    \begin{tabular}{lcccccc}
        \toprule
        \textbf{Model} & \textbf{Res.} & \textbf{\# of}     & \textbf{FLOPs} & \textbf{Thru.} & \textbf{Memory} & \textbf{Top-1}\\
                       &               & \textbf{Params}    &                & (imgs/sec)     & (GB)            & (\%) \\
        \midrule
        \multicolumn{7}{c}{\textit{ImageNet-1K trained models}}\\
        \midrule
        \nb\textbf{NAT-M}                           & 224\sq &  20 M &   2.7 G & 2132 &  2.4 & \textbf{81.8} \\
        \ours \textbf{DiNAT-M}                      & 224\sq &  20 M &   2.7 G & 2080 &  2.4 & \textbf{81.8} \\
        \midrule
        \wb\textbf{Swin-T}                          & 224\sq &  28 M &   4.5 G & 1724 &  4.8 & 81.3 \\
        \ours \textbf{DiNAT$_s$-T}                  & 224\sq &  28 M &   4.5 G & 1954 &  4.0 & 81.8 \\
        \cb\textbf{ConvNeXt-T}                      & 224\sq &  28 M &   4.5 G & 2491 &  3.4 & 82.1 \\
        \nb\textbf{NAT-T}                           & 224\sq &  28 M &   4.3 G & 1537 &  2.5 & \textbf{83.2} \\
        \ours \textbf{DiNAT-T}                      & 224\sq &  28 M &   4.3 G & 1500 &  2.5 & 82.7 \\
        \midrule
        \wb\textbf{Swin-S}                          & 224\sq &  50 M &   8.7 G & 1056 &  5.0 & 83.0 \\
        \ours \textbf{DiNAT$_s$-S}                  & 224\sq &  50 M &   8.7 G & 1203 &  4.1 & 83.5 \\
        \cb\textbf{ConvNeXt-S}                      & 224\sq &  50 M &   8.7 G & 1549 &  3.5 & 83.1 \\
        \nb\textbf{NAT-S}                           & 224\sq &  51 M &   7.8 G & 1049 &  3.7 & 83.7 \\
        \ours \textbf{DiNAT-S}                      & 224\sq &  51 M &   7.8 G & 1058 &  3.7 & \textbf{83.8} \\
        \midrule
        \wb\textbf{Swin-B}                          & 224\sq &  88 M &  15.4 G &  774 &  6.7 & 83.5 \\
        \ours \textbf{DiNAT$_s$-B}                  & 224\sq &  88 M &  15.4 G &  877 &  5.5 & 83.8 \\
        \cb\textbf{ConvNeXt-B}                      & 224\sq &  89 M &  15.4 G & 1107 &  4.8 & 83.8 \\
        \nb\textbf{NAT-B}                           & 224\sq &  90 M &  13.7 G &  781 &  5.0 & 84.3 \\
        \ours \textbf{DiNAT-B}                      & 224\sq &  90 M &  13.7 G &  764 &  5.0 & \textbf{84.4} \\
        \midrule
        \multicolumn{7}{c}{\textit{ImageNet-22K pre-trained models}}\\
        \midrule
        \wb\textbf{Swin-L}                          & 224\sq & 197 M &  34.5 G &  478 & 10.4 & 86.3 \\
        \ours \textbf{DiNAT$_s$-L}                  & 224\sq & 197 M &  34.5 G &  528 &  8.6 & 86.5 \\
        \cb\textbf{ConvNeXt-L}                      & 224\sq & 198 M &  34.4 G &  643 &  7.5 & \textbf{86.6} \\
        \ours \textbf{DiNAT-L}                      & 224\sq & 200 M &  30.6 G &  474 &  7.8 & \textbf{86.6} \\
        \midrule
        \wb\textbf{Swin-L\dgr}                      & 384\sq & 197 M & 104.0 G &  169 & 32.7 & 87.3 \\
        \ours \textbf{DiNAT$_s$-L}                  & 384\sq & 197 M & 101.5 G &  181 & 22.6 & 87.4 \\
        \cb\textbf{ConvNeXt-L}                      & 384\sq & 198 M & 101.1 G &  221 & 19.2 & \textbf{87.5} \\
        \ours \textbf{DiNAT-L}                      & 384\sq & 200 M &  89.7 G &  161 & 20.1 & 87.4 \\
        \ours \textbf{DiNAT-L\dgr}                  & 384\sq & 200 M &  92.4 G &  110 & 26.9 & \textbf{87.5} \\
        \bottomrule
    \end{tabular}
    }
    \caption{
    \textbf{ImageNet-1K image classification performance.} 
    \dgr indicates increased window size from 7\sq{} to 11\sq{} (DiNAT) and 12\sq{} (Swin). 
    Throughput and peak memory usage are measured from forward passes with a batch size of 256 on a single A100 GPU.
    Note that DiNAT$_s$ is identical in architecture to Swin, and only different in attention modules (WSA/SWSA replaced with NA/DiNA).
    }
    \label{apptab:imagenet_comparison}
\end{table}

\setlength{\tabcolsep}{3pt}
\begin{table}[t]
    \centering
    \resizebox{0.475\textwidth}{!}{
    \begin{tabular}{lccc|ccc|ccc}
        \toprule
        \textbf{Backbone} & \textbf{\# of} & \textbf{FLOPs} & \textbf{Thru.} & \textbf{AP\textsuperscript{b}} & \textbf{AP\textsuperscript{b}\textsubscript{50}} & \textbf{AP\textsuperscript{b}\textsubscript{75}} & \textbf{AP\textsuperscript{m}} & \textbf{AP\textsuperscript{m}\textsubscript{50}} & \textbf{AP\textsuperscript{m}\textsubscript{75}} \\ 
        & \textbf{Params} &&(FPS)\\
        \midrule
        \multicolumn{10}{c}{\textit{Mask R-CNN - 3x schedule}} \\
        \midrule
        \nb\textbf{NAT-M}                                       &  40 M &  225 G & 54.1 & 46.5 & 68.1 & 51.3 & 41.7 & 65.2 & 44.7 \\
        \ours \textbf{DiNAT-M}                                  &  40 M &  225 G & 53.8 & \textbf{47.2} & \textbf{69.1} & \textbf{51.9} & \textbf{42.5} & \textbf{66.0} & \textbf{45.9} \\
        \midrule
        \wb\textbf{Swin-T}                                      &  48 M &  267 G & 45.1 & 46.0 & 68.1 & 50.3 & 41.6 & 65.1 & 44.9 \\
        \ours \textbf{DiNAT$_s$-T}                              &  48 M &  263 G & 52.5 & 46.6 & 68.8 & 51.3 & 42.1 & 65.7 & 45.4 \\
        \cb\textbf{ConvNeXt-T}                                  &  48 M &  262 G & 52.0 & 46.2 & 67.0 & 50.8 & 41.7 & 65.0 & 44.9 \\
        \nb\textbf{NAT-T}                                       &  48 M &  258 G & 44.5 & 47.7 & 69.0 & 52.6 & 42.6 & 66.1 & 45.9 \\
        \ours \textbf{DiNAT-T}                                  &  48 M &  258 G & 43.3 & \textbf{48.6} & \textbf{70.2} & \textbf{53.4} & \textbf{43.5} & \textbf{67.3} & \textbf{46.8} \\
        \midrule
        \wb\textbf{Swin-S}                                      &  69 M &  359 G & 31.7 & 48.5 & 70.2 & 53.5 & 43.3 & 67.3 & 46.6 \\
        \ours \textbf{DiNAT$_s$-S}                              &  69 M &  350 G & 38.7 & 48.6 & 70.4 & 53.2 & 43.5 & 67.6 & 46.9 \\
        \nb\textbf{NAT-S}                                       &  70 M &  330 G & 34.8 & 48.4 & 69.8 & 53.2 & 43.2 & 66.9 & 46.5 \\
        \ours \textbf{DiNAT-S}                                  &  70 M &  330 G & 35.3 & \textbf{49.3} & \textbf{70.8} & \textbf{54.2} & \textbf{44.0} & \textbf{68.0} & \textbf{47.4} \\
        \midrule
        \multicolumn{10}{c}{\textit{Cascade Mask R-CNN - 3x schedule}} \\
        \midrule
        \nb\textbf{NAT-M}                                       &  77 M &  704 G & 27.8 & 50.3 & 68.9 & 54.9 & 43.6 & 66.4 & 47.2 \\
        \ours \textbf{DiNAT-M}                                  &  77 M &  704 G & 27.6 & \textbf{51.2} & \textbf{69.8} & \textbf{55.7} & \textbf{44.4} & \textbf{67.3} & \textbf{47.8} \\
        \midrule
        \wb\textbf{Swin-T}                                      &  86 M &  745 G & 25.1 & 50.4 & 69.2 & 54.7 & 43.7 & 66.6 & 47.3 \\
        \ours \textbf{DiNAT$_s$-T}                              &  86 M &  742 G & 27.4 & 51.0 & 69.9 & 55.4 & 44.1 & 67.3 & 47.6 \\
        \cb\textbf{ConvNeXt-T}                                  &  86 M &  741 G & 27.3 & 50.4 & 69.1 & 54.8 & 43.7 & 66.5 & 47.3 \\
        \nb\textbf{NAT-T}                                       &  85 M &  737 G & 24.9 & 51.4 & 70.0 & 55.9 & 44.5 & 67.6 & 47.9 \\
        \ours \textbf{DiNAT-T}                                  &  85 M &  737 G & 25.0 & \textbf{52.2} & \textbf{71.0} & \textbf{56.8} & \textbf{45.1} & \textbf{68.3} & \textbf{48.8} \\
        \midrule
        \wb\textbf{Swin-S}                                      & 107 M &  838 G & 20.3 & 51.8 & 70.4 & 56.3 & 44.7 & 67.9 & 48.5 \\
        \ours \textbf{DiNAT$_s$-S}                              & 107 M &  829 G & 23.1 & 52.3 & 71.2 & 56.7 & 45.2 & 68.6 & 49.1 \\
        \cb\textbf{ConvNeXt-S}                                  & 108 M &  827 G & 23.0 & 51.9 & 70.8 & 56.5 & 45.0 & 68.4 & 49.1 \\
        \nb\textbf{NAT-S}                                       & 108 M &  809 G & 21.7 & 52.0 & 70.4 & 56.3 & 44.9 & 68.1 & 48.6 \\
        \ours \textbf{DiNAT-S}                                  & 108 M &  809 G & 21.8 & \textbf{52.9} & \textbf{71.8} & \textbf{57.6} & \textbf{45.8} & \textbf{69.3} & \textbf{49.9} \\
        \midrule
        \wb\textbf{Swin-B}                                      & 145 M &  982 G & 17.3 & 51.9 & 70.9 & 56.5 & 45.0 & 68.4 & 48.7 \\
        \ours \textbf{DiNAT$_s$-B}                              & 145 M &  966 G & 19.7 & 52.6 & 71.5 & 57.2 & 45.3 & 68.8 & 49.1 \\
        \cb\textbf{ConvNeXt-B}                                  & 146 M &  964 G & 19.5 & 52.7 & 71.3 & 57.2 & 45.6 & 68.9 & 49.5 \\
        \nb\textbf{NAT-B}                                       & 147 M &  931 G & 18.6 & 52.3 & 70.9 & 56.9 & 45.1 & 68.3 & 49.1 \\
        \ours \textbf{DiNAT-B}                                  & 147 M &  931 G & 18.5 & \textbf{53.4} & \textbf{72.1} & \textbf{58.2} & \textbf{46.2} & \textbf{69.7} & \textbf{50.2} \\
        \midrule
        \wb\textbf{Swin-L\strr\dgg}                             & 253 M & 1393 G & 12.9 & 53.7 & 72.2 & 58.7 & 46.4 & 69.9 & 50.7 \\
        \ours \textbf{DiNAT$_s$-L\dgg}                          & 253 M & 1357 G & 15.0 & 54.8 & 74.2 & 59.8 & 47.2 & 71.3 & 51.2 \\
        \cb\textbf{ConvNeXt-L\dgg}                              & 253 M & 1354 G & 14.8 & 54.8 & 73.8 & 59.8 & 47.6 & 71.3 & 51.7 \\
        \ours \textbf{DiNAT-L\dgg}                              & 258 M & 1276 G & 14.0 & \textbf{55.3} & \textbf{74.3} & \textbf{60.2} & \textbf{47.8} & \textbf{71.8} & \textbf{52.0} \\
        \bottomrule
    \end{tabular}
    }
    \caption{
    \textbf{COCO object detection and instance segmentation performance.} 
    \dgg indicates that the model was pre-trained on ImageNet-22K. 
    \strr Swin-L was not reported with Cascade Mask R-CNN, therefore we trained it with their official checkpoint. 
    Throughput is measured on a single A100 GPU.
    Note that DiNAT$_s$ is identical in architecture to Swin, and only different in attention modules (WSA/SWSA replaced with NA/DiNA).
    }
    \label{apptab:objectdetection}
\end{table}

\setlength{\tabcolsep}{5pt}
\begin{table}[t]
    \centering
    \resizebox{0.475\textwidth}{!}{
    \begin{tabular}{lc|c|cc|c}
        \toprule
        \textbf{Model}                              & \textbf{Dilation}             & \textbf{ImageNet} & \multicolumn{2}{c}{\textbf{MSCOCO}}                               & \textbf{ADE20K}   \\
                                                    & \textbf{per level}            & \textbf{Top-1 (\%)}& \textbf{AP\textsuperscript{b}} & \textbf{AP\textsuperscript{m}}   & \textbf{mIoU}     \\
        \midrule
        \wb\textbf{Swin-Tiny}                       & Not Applicable                & 81.3    & 46.0                           & 41.6                             & 45.8              \\
        \nb\textbf{NAT$_s$-Tiny}                    & 1, 1, 1, 1                    & 81.8    & 46.1                           & 41.5                             & 46.2              \\
        \db\textbf{DiNAT$_s$-Tiny}                  & 8, 4, 2, 1                    & 81.8    & 46.3                           & 41.6                             & 46.7              \\
        \db\textbf{DiNAT$_s$-Tiny}                  & 16, 8, 4, 2                   & -       & \textbf{46.4}                  & \textbf{41.8}                    & 47.1              \\
        \db\textbf{DiNAT$_s$-Tiny}                  & Maximum                       & 81.8    & \textbf{46.4}                  & \textbf{41.9}                    & 47.0              \\
        \ours \textbf{DiNAT$_s$-Tiny}               & Gradual                       & -       & \textbf{46.6}                  & \textbf{42.1}                    & \textbf{47.4}     \\
        \midrule
        \nb\textbf{NAT-Tiny}                        & 1, 1, 1, 1                    & 83.2    & 47.7                           & 42.6                             & 48.4              \\
        \db\textbf{DiNAT-Tiny}                      & 8, 4, 2, 1                    & 82.7    & 48.0                           & 42.9                             & 48.5              \\
        \db\textbf{DiNAT-Tiny}                      & 16, 8, 4, 2                   & -       & 48.3                           & 43.4                             & 48.5              \\
        \db\textbf{DiNAT-Tiny}                      & Maximum                       & 82.7    & \textbf{48.6}                  & \textbf{43.5}                    & 48.7              \\
        \ours \textbf{DiNAT-Tiny}                   & Gradual                       & -       & \textbf{48.6}                  & \textbf{43.5}                    & \textbf{48.8}     \\
        \bottomrule
    \end{tabular}
    }
    \caption{
    \textbf{Dilation impact on performance.} 
    Models listed within the same section have identical architectures and are different only in attention patterns (NAT$_s$ is identical to Swin with both WSA and SWSA replaced with NA, DiNAT$_s$ replaces SWSA with DiNA).
    }
    \label{apptab:dilationperformance}
\end{table}

\setlength{\tabcolsep}{3pt}
\begin{table}[t]
    \centering
    \resizebox{0.475\textwidth}{!}{
    \begin{tabular}{l|cccc|cc}
        \toprule
        \textbf{Backbone} & \textbf{Res.} & \textbf{\# of}     & \textbf{FLOPs} & \textbf{Thru.} & \multicolumn{2}{c}{\textbf{mIoU}}\\ 
                          &               & \textbf{Params}    &                & (FPS)          & single scale & multi scale \\ 
        \midrule
        \nb\textbf{NAT-M}                           & 2048 \texttimes{} 512 &  50 M &  900 G & 24.5 & 45.1 & 46.4 \\
        \ours \textbf{DiNAT-M}                      & 2048 \texttimes{} 512 &  50 M &  900 G & 24.2 & \textbf{45.8} & \textbf{47.2} \\
        \midrule
        \wb\textbf{Swin-T}                          & 2048 \texttimes{} 512 &  60 M &  946 G & 21.3 & 44.5 & 45.8 \\
        \ours \textbf{DiNAT$_s$-T}                  & 2048 \texttimes{} 512 &  60 M &  941 G & 23.5 & 46.0 & 47.4 \\
        \cb\textbf{ConvNeXt-T}                      & 2048 \texttimes{} 512 &  60 M &  939 G & 23.3 & 46.0 & 46.7 \\
        \nb\textbf{NAT-T}                           & 2048 \texttimes{} 512 &  58 M &  934 G & 21.4 & 47.1 & 48.4 \\
        \ours \textbf{DiNAT-T}                      & 2048 \texttimes{} 512 &  58 M &  934 G & 21.3 & \textbf{47.8} & \textbf{48.8} \\
        \midrule
        \wb\textbf{Swin-S}                          & 2048 \texttimes{} 512 &  81 M & 1040 G & 17.0 & 47.6 & 49.5 \\
        \ours \textbf{DiNAT$_s$-S}                  & 2048 \texttimes{} 512 &  81 M & 1030 G & 19.1 & 48.6 & \textbf{49.9} \\
        \cb\textbf{ConvNeXt-S}                      & 2048 \texttimes{} 512 &  82 M & 1027 G & 19.1 & 48.7 & 49.6 \\
        \nb\textbf{NAT-S}                           & 2048 \texttimes{} 512 &  82 M & 1010 G & 17.9 & 48.0 & 49.5 \\
        \ours \textbf{DiNAT-S}                      & 2048 \texttimes{} 512 &  82 M & 1010 G & 18.1 & \textbf{48.9} & \textbf{49.9} \\
        \midrule
        \wb\textbf{Swin-B}                          & 2048 \texttimes{} 512 & 121 M & 1188 G & 14.6 & 48.1 & 49.7 \\
        \ours \textbf{DiNAT$_s$-B}                  & 2048 \texttimes{} 512 & 121 M & 1173 G & 16.5 & 49.4 & 50.2 \\
        \cb\textbf{ConvNeXt-B}                      & 2048 \texttimes{} 512 & 122 M & 1170 G & 16.4 & 49.1 & 49.9 \\
        \nb\textbf{NAT-B}                           & 2048 \texttimes{} 512 & 123 M & 1137 G & 15.6 & 48.5 & 49.7 \\
        \ours \textbf{DiNAT-B}                      & 2048 \texttimes{} 512 & 123 M & 1137 G & 15.4 & \textbf{49.6} & \textbf{50.4} \\
        \midrule
        \wb\textbf{Swin-L\dgr\dgg}                  & 2560 \texttimes{} 640 & 234 M & 2585 G &  8.5 &   -  & 53.5 \\
        \ours \textbf{DiNAT$_s$-L\dgg}              & 2560 \texttimes{} 640 & 234 M & 2466 G &  9.7 & 53.4 & 54.6 \\
        \cb\textbf{ConvNeXt-L\dgg}                  & 2560 \texttimes{} 640 & 235 M & 2458 G &  9.6 & 53.2 & 53.7 \\
        \ours \textbf{DiNAT-L\dgg}                  & 2560 \texttimes{} 640 & 238 M & 2335 G &  9.0 & \textbf{54.0} & \textbf{54.9} \\
        \bottomrule
    \end{tabular}
    }
    \caption{
    \textbf{ADE20K semantic segmentation performance.} 
    \dgg indicates that the model was pre-trained on ImageNet-22K. 
    \dgr indicates increased window size from 7\sq{} to 12\sq{}. 
    Throughput is measured on a single A100 GPU.
    Note that DiNAT$_s$ is identical in architecture to Swin, and only different in attention modules (WSA/SWSA replaced with NA/DiNA).
    }
    \label{apptab:semseg}
\end{table}

In \cref{subsec:miscexps} we experimented with architecture-related hyperparameters that are introduced by DiNA: dilation values, and the ordering of NA and DiNA layers.
We also complete those dilation experiments by adding DiNAT$_s$ and Swin, and presente the results in \cref{apptab:dilationperformance,apptab:nadinaorderperformance}.
\setlength{\tabcolsep}{5pt}
\begin{table}[t]
    \centering
    \resizebox{0.475\textwidth}{!}{
    \begin{tabular}{lc|c|cc|c}
        \toprule
        \textbf{Variant}                                & \textbf{Layer}            & \textbf{ImageNet} & \multicolumn{2}{c|}{\textbf{MSCOCO}}                              & \textbf{ADE20K}   \\
                                                        & \textbf{structure}        & \textbf{Top-1 (\%)}& \textbf{AP\textsuperscript{b}} & \textbf{AP\textsuperscript{m}}   & \textbf{mIoU}     \\
        \midrule
        \wb\wb\textbf{Swin-Tiny}                        & WSA-SWSA                  & 81.3            & 46.0                           & 41.6                             & 45.8              \\
        \nb\nb\textbf{NAT$_s$-Tiny}                     & NA-NA                     & \textbf{81.8}   & 46.1                           & 41.5                             & 46.2              \\
        \grayrow\nb\db \textbf{DiNAT$_s$-Tiny}          & NA-DiNA                   & \textbf{81.8}   & 46.4                           & \textbf{41.8}                    & \textbf{47.1}     \\
        \db\nb                                          & DiNA-NA                   & 81.5            & \textbf{46.5}                  & \textbf{41.8}                    & 46.9              \\
        \db\db                                          & DiNA-DiNA                 & 79.7            & 39.8                           & 36.8                             & 40.7              \\
        \midrule
        \nb\nb\textbf{NAT-Tiny}                         & NA-NA                     & \textbf{83.2}   & 47.7                           & 42.6                             & 48.4              \\
        \grayrow\nb\db\textbf{DiNAT-Tiny}               & NA-DiNA                   & 82.7            & 48.3                           & 43.4                             & \textbf{48.5}     \\
        \db\nb                                          & DiNA-NA                   & 82.6            & \textbf{48.5}                  & \textbf{43.5}                    & 47.9              \\
        \db\db                                          & DiNA-DiNA                 & 82.2            & 44.9                           & 40.5                             & 45.8              \\
        \bottomrule
    \end{tabular}
    }
    \caption{
    \textbf{Layer structure impact on performance.} 
    }
    \label{apptab:nadinaorderperformance}
\end{table}

\end{document}